\documentclass[11pt]{amsart}
\usepackage{amsfonts,amsmath,amssymb,amsthm}
\usepackage{mathtools}
\mathtoolsset{showonlyrefs}
\usepackage{booktabs}
\usepackage{graphicx}
\usepackage[margin=1.1in]{geometry}
\usepackage[hidelinks]{hyperref}
\usepackage{microtype}

\usepackage{xcolor}
\usepackage{verbatim}
\allowdisplaybreaks
\hypersetup{
  pdftitle={Deep Neural Operator Learning for Semilinear SPDEs},
  pdfauthor={Hengrong Du, Qi Feng, and Weinan Wang},
  pdfsubject={Finite Fourier measurements, Schauder stability, and operator approximation},
  pdfkeywords={deep operator networks, neural operators, universal approximation,
    semilinear stochastic PDEs, Gaussian forcing, parabolic Holder-Besov spaces,
    Fourier measurements, Wiener chaos, stochastic Navier-Stokes equations}
}
 
\newtheorem{theorem}{Theorem}[section]
\newtheorem{lemma}[theorem]{Lemma}
\newtheorem{proposition}[theorem]{Proposition}
\newtheorem{corollary}[theorem]{Corollary}
\newtheorem{definition}[theorem]{Definition}
\newtheorem{assumption}[theorem]{Assumption}
\newtheorem{remark}[theorem]{Remark}
 
\newcommand{\cC}{\mathcal C}
\newcommand{\cG}{\mathcal G}
\newcommand{\cS}{\mathcal S}
\newcommand{\R}{\mathbb R}
\newcommand{\T}{\mathbb T}
\newcommand{\Z}{\mathbb Z}
\newcommand{\E}{\mathbb E}
\newcommand{\PP}{\mathbb P}
\newcommand{\one}{\mathbf 1}
\newcommand{\DT}{D_T}
\newcommand{\Rec}{\operatorname{Rec}}
\newcommand{\FNN}{\operatorname{FNN}}
 
\title{Deep Neural Operator Learning for stochastic heat equations and applications}
\author{Hengrong Du}
\address{Department of Mathematics and Computer Science, Fisk University, Nashville, TN 37208, USA}
\email{hdu@fisk.edu}
\author{Qi Feng}
\address{Department of Mathematics, Florida State University, Tallahassee, FL 32306, USA}
\email{qfeng2@fsu.edu}
\author{Weinan Wang}
\address{Department of Mathematics, University of Oklahoma, Norman, OK 73019, USA}
\email{ww@ou.edu}
\thanks{QF is partially supported by the National Science Foundation under grant \#DMS-2420029.
WW was partially supported by the Simons Foundation Travel Support for Mathematicians (No. 0007730).}
\date{}

\begin{document}

\begin{abstract}
In this paper, we establish a universal approximation theorem for solution operators of semilinear stochastic heat equations in spatial dimension \(d\ge2\), extending deep neural operator approximation results to SPDEs with distribution-valued Gaussian forcing. A smooth Fourier cutoff reduces the forcing to finitely many measurements, while Besov and Schauder estimates control the resulting solution error. The finite-dimensional solution map is then approximated by ReLU branch networks together with grid-based trunk functions, yielding explicit bounds on the numbers of forcing measurements and trunk functions and approximation in every finite stochastic moment. For linear equations, the reduction coincides with truncation of the first Wiener chaos. Lastly, as an application, we prove a high-probability local-in-time approximation theorem for stochastic Navier--Stokes equations, with finite measurements of both the initial condition and forcing and an explicit sensing rate. 
\end{abstract}
\maketitle

\section{Introduction}

Operator learning seeks a single approximation of the map from the data of
an equation to its solution, enabling evaluation at new inputs without
solving each instance input. The operator approximation theorems in~\cite{chen1995}, DeepONet~\cite{lu2021deeponet}, and Fourier neural
operators~\cite{li2021fno,kovachki2023neural} have established such approximation for maps between function spaces. 
For stochastic partial differential equations (SPDEs), we study solution operators that map initial data and random forcing to the corresponding solutions. The central challenge is that the forcing may be distribution-valued, requiring an operator approximation framework that accommodates inputs without pointwise values.

This paper develops a finite-measurement approximation theory for such SPDE
solution operators. Our starting point is the probabilistic operator
framework in~\cite{bayraktar2025deep},
which combines integrability, tail estimates, and Lipschitz continuity for
models driven by finite-dimensional stochastic processes. Extending this
approach to distribution-valued space--time forcing requires an encoding
that is well defined on the noise space and an estimate that propagates its
error through the equation. We use Fourier measurements to construct the encoding and Schauder stability to control the resulting error. The resulting branch--trunk
network approximates the joint dependence on the initial condition and the
forcing in stochastic moments of the uniform space--time norm, with
explicit measurement and reconstruction rates.

We consider the semilinear stochastic heat equation
\begin{equation}\label{eq:she}
\begin{cases}
    \partial_t u-\Delta u=F(u)+\eta,
       &(t,x)\in(0,T]\times\T^d,\\
    u(0,x)=u_0(x),
       &x\in\T^d,
\end{cases}
\end{equation}
where $d\ge2$, $F:\R\to\R$ is globally Lipschitz,
$u_0\in C(\T^d)$, and $\eta$ is a centered Gaussian
random distribution (as defined in Assumption~\ref{ass:noise} below).
The equation is interpreted pathwisely in the mild sense.
For this SPDE, our main result
constructs ReLU networks of the form
\begin{equation}\label{eq:intro-deeponet}
    \Gamma_\theta(u_0,\eta)(t,x)
    =\sum_{\ell=1}^{N_1}
       a_\ell\big(\vec u_0,c_J(\eta)\big)q_\ell(t,x),
\end{equation}
where $\vec u_0$ contains $N_0$ point values of the initial condition and
$c_J(\eta)$ contains $N_J$ real continuous linear measurements of the
forcing. For every fixed $q\in[1,\infty)$ and
$\varepsilon\in(0,1)$, one such network satisfies
\begin{equation}\label{eq:intro-moment-bound}
    \sup_{u_0\in\cG_0}
    \left(\E\left[
       \|\cS(u_0,\eta)-\Gamma_\theta(u_0,\eta)\|_{C(D_T)}^q
    \right]\right)^{1/q}
    \le\varepsilon.
\end{equation}
Thus the approximation controls full solution paths, uniformly over the
initial-data class, while retaining the prescribed coupling between the
noise input and the solution output. 
The main analytic step is a quantitative reduction of the rough/noisy input.
A smooth finite-rank Fourier cutoff satisfies
\[
    \|(I-\Pi_J)\eta\|_{\cC^{\beta_1}}
    \lesssim 2^{-J\delta}\|\eta\|_{\cC^{\beta_0}}.
\]
The main difficulty is to turn this finite-dimensional reduction into an approximation of the SPDE solution operator. The forcing may not have pointwise values, and its truncation error must be measured in a topology in which the solution map is stable.
We combine the stability estimate for \eqref{eq:she} with partition-of-unity approximation of the finite-dimensional input dependence and parabolic interpolation of the output. This extends the function, functional, and operator approximation
approach in~\cite{liu2024neural,bayraktar2025deep} to
distribution-valued forcing, with explicit bounds on the numbers
of input measurements and trunk functions.

Another difficulty is that the Gaussian input is not confined to a compact set. Therefore Uniform approximation on bounded sets alone does not yield the stochastic moment estimate above. We control the growth of the neural approximation outside these sets by a coordinatewise projection, and use Fernique’s theorem to estimate the remaining Gaussian tail. The construction consequently gives approximation in every fixed finite stochastic moment without imposing compact support on the noise law.

For stochastic incompressible Navier–Stokes equations, the quadratic transport term contains a spatial derivative, and the solution map is only locally Lipschitz. We establish a common deterministic existence interval for the admissible initial velocities and forcings in bounded sets, together with their Fourier truncations. On this interval, which is independent of the approximation accuracy, local stability yields uniform operator approximation on an event of high probability. The measurement bounds distinguish the regularity required of the initial velocity from that of the forcing.

\subsection{Related literature}

The deterministic operator-learning literature supplies the approximation
architecture used below.  Alongside DeepONet, the Fourier neural operator
in~\cite{li2021fno} and the general neural-operator framework
in~\cite{kovachki2023neural} provide resolution-independent
parameterizations of maps between function spaces.  Universal approximation
and error estimates for Fourier neural operators are proved in
\cite{kovachki2021fno}; the encoding--approximation--reconstruction analysis
of DeepONets in \cite{lanthaler2022error} is particularly close to the error
decomposition used here.  Random-feature operator learning \cite{nelsen2024operator} gives a
probabilistic alternative , and its statistical error
analysis extends to infinite-dimensional, vector-valued regression
\cite{lanthaler2023vector}.  Learning linear operators from noisy input--output
pairs is studied in \cite{dehoop2023convergence}.  These general results also
show why quantitative branch-network bounds are difficult without additional
structure: \emph{a generic Lipschitz operator on a high-dimensional encoded input
class retains the curse of dimensionality}.

Operator learning has also been extended to geometry-dependent PDE
solution maps. The DIMON framework~\cite{yin2024geometry} transports problems
on a family of diffeomorphic domains to a common reference domain,
where the dependence on the geometry and initial or boundary data is
learned.

Finite-dimensional inputs and outputs also arise only when  parameters or
observables of an underlying PDE solution are available.  Fourier neural
mappings for such parameter-to-observable maps are developed in
\cite{huang2025parameter}.  In the inverse direction, neural inverse operators
combine DeepONet and Fourier-neural-operator components to learn PDE inverse
maps \cite{molinaro2023inverse}; see \cite{nelsen2026inverse} for a broader
probabilistic account of operator learning for inverse problems.  Those works
concern statistical recovery from observations, whereas the present paper
constructs a pathwise approximation of a forward SPDE solution map after
encoding its distribution-valued forcing.

Operator learning for stochastic models has applications in mathematical
finance, control, and game theory. Solution operators for structured
families of forward--backward stochastic differential equations with
random terminal time are approximated in \cite{furuya2024simultaneously}.
In dynamic Stackelberg games, attention-based neural operators approximate
the follower's best-response operator \cite{alvarez2024neural}. Neural
operators for families of convex optimization problems are developed
in \cite{kratsios2025convex}, with applications to stochastic optimal
control, portfolio optimization, and hedging. These operator
formulations complement methods such as the Deep-BSDE method
\cite{han2018deep}, which approximates the solution of a fixed equation. For optimal execution problems with unknown price impact,
the authors of \cite{meng2025execution} combine in-context learning of
the price-impact operator from observed trade and impact trajectories
with neural-network optimization of the execution policy. Another interesting in-context operator learning on spaces of probability
measures is studied in~\cite{cole2026incontext}, using samples
from source and target distributions to approximate optimal transport
maps.

The authors of \cite{kummerle2024transition}
study the recovery of a discrete-time
linear transition operator from sparse observations of its powers.
In \cite{lu2022stochastic}, the authors establish consistency
and learning rates for interaction kernels inferred from multiple
trajectories of stochastic particle systems. These works concern
inference of the governing dynamics, whereas one can fix the SDE and
approximate its solution operator as a function of the initial
condition and distribution-valued forcing. In \cite{an2025autobalance},
the authors apply separate adaptive optimizers to
the individual loss terms before combining their updates.
Curvature-aware corrections to first-order optimizers are developed
in \cite{an2026lightweight}, while a regularized quasi-Newton method
for high-accuracy training is proposed in \cite{si2026score}.
These works address training algorithms; our results concern the
existence and size of approximating networks.

Several architectures have been designed specifically for stochastic PDEs.
Neural SPDEs \cite{salvi2022neural} build the mild-solution structure into a
resolution-invariant model.  Feature engineering with regularity structures
\cite{chevyrev2024features}, NORS \cite{hu2022nors}, and the deep latent
regularity network \cite{gong2023deep} use polynomial or renormalized features
of the driving noise.  Such features are indispensable for singular equations
in which the solution and its nonlinearities are not classically defined.
SPDEBench \cite{spdebench2025} provides a numerical comparison across regular
and singular stochastic equations.  The present paper addresses a different
question: \emph{for additive Gaussian forcing whose heat convolution is continuous,
are linear measurements of the distributional input already sufficient for a
pathwise operator-approximation theorem?}  The answer is affirmative under
Assumption~\ref{ass:noise}.

Navier--Stokes solution operators have also been used as important test cases
for neural operators.  Quantitative approximation results for deterministic
incompressible Navier--Stokes appear in \cite{kovachki2021fno}, while Neural
SPDEs \cite{salvi2022neural} and NORS \cite{hu2022nors} include stochastic
Navier--Stokes experiments. In \cite{wang2026conformal}, the authors introduced a conformal prediction approach from statistics to study the operator learning of deterministic Navier-Stokes equations.  Our result is complementary to these works: We prove that finitely many linear measurements of the forcing and initial velocity suffice for uniform approximation on a common local existence interval, on an event of high probability.  The measurement counts in
Theorem \ref{thm:nse-approx} separate the forcing regularity from the initial
velocity regularity.

The finite measurements also connect the construction to Wiener chaos.
Classical chaos solutions of stochastic evolution equations are developed in
\cite{mikulevicius1998chaos,lototsky2006chaos}; recent work learns the
deterministic propagators in truncated chaos expansions
\cite{neufeld2024chaos,shi2026chaos,shi2026singular}.  The precise relation is
recorded in Lemma \ref{lem:wce-linear}: for the linear equation, the
finite-rank solution used here is exactly the truncation of the first-chaos expansion to the retained coordinates.  For a
general globally Lipschitz nonlinearity, the present proof instead uses
Schauder stability and does not require a closed propagator hierarchy.  Our approximation argument follows the function, functional, and operator approximation results of \cite{bayraktar2025deep}.  Deep-BSDE methods target the solution of a fixed
equation, whereas the present branch--trunk network approximates the joint
dependence of the solution on the initial condition and the forcing
realization.

The paper is organized as follows.  Section \ref{sec:setting} states preliminaries, the
analytic assumptions, and proves well-posedness and Lipschitz stability.
Section \ref{sec:reduction} constructs the finite-dimensional measurement map and the
deep operator architecture.  Section \ref{sec:approximation} proves the
function, functional, and operator approximation results.
Section \ref{sec:nse} develops the
high-probability local extension to stochastic Navier--Stokes equations.

\section{Preliminaries}\label{sec:setting}
In this section, we specify the function spaces and give some preliminaries.

\subsection{Fourier multipliers and function spaces}\label{sec:pre-operators}

Fix $T>0$ and write
\[
    \T^d=(\R/2\pi\Z)^d,\qquad
    \T_T=\R/T\Z,\qquad
    \Lambda_T=\T_T\times\T^d,\qquad
    \DT=[0,T]\times\T^d.
\]
The forcing is realized on the space--time torus $\Lambda_T$ and then
restricted to the causal equation on $\DT$. The solution is not
required to be periodic in time. On $\DT$ we use the parabolic distance
\[
    d_{\mathfrak s}((t,x),(s,y))
       :=|t-s|^{1/2}+d_{\T^d}(x,y),
    \qquad
    d_{\T^d}(x,y):=\min_{m\in\Z^d}|x-y-2\pi m|.
\]
This distance corresponds to the scaling
$\mathfrak s=(2,1,\ldots,1)$, whose dimension is
$|\mathfrak s|=d+2$. For $0<\gamma<1$, we write
\[
    \|f\|_{C^\gamma_{\mathfrak s}(\DT)}
       :=\|f\|_\infty+
          \sup_{z\ne z'}\frac{|f(z)-f(z')|}
                             {d_{\mathfrak s}(z,z')^\gamma}.
\]
All function-space norms below are deterministic norms. Expectations
are taken separately when these norms are applied to random functions.
For vector-valued functions, we take the maximum over components.

The orthonormal Fourier basis on $\Lambda_T$ are
\begin{equation}\label{eq:pre-characters}
    \varepsilon_{n,k}(t,x)
       :=\big(T(2\pi)^d\big)^{-1/2}
          e^{2\pi int/T+ik\cdot x},
    \qquad (n,k)\in\Z\times\Z^d.
\end{equation}
For a distribution $f$, its Fourier coefficient $\widehat f(n,k)$ is
its action on $\overline{\varepsilon_{n,k}}$, i.e., $\widehat f(n,k)=\langle f, \overline{\varepsilon_{n,k}}\rangle$. The corresponding
spatial characters are $e_k(x)=(2\pi)^{-d/2}e^{ik\cdot x}$.

Now we fix a smooth dyadic decomposition compatible with parabolic
scaling: choose a nonincreasing function
$\varsigma\in C^\infty([0,\infty);[0,1])$ that equals one on $[0,1]$
and vanishes on $[4/3,\infty)$. Set
\[
    \varrho(\omega,\kappa)
       :=(\omega^2+|\kappa|^4)^{1/4},
    \qquad
    \chi(\omega,\kappa):=\varsigma(\varrho(\omega,\kappa)).
\]
Although $\varrho$ is not smooth at the origin, $\chi$ is smooth there
because it is constant in a neighborhood of the origin. Moreover,
$\varrho(\lambda^2\omega,\lambda\kappa)=\lambda\varrho(\omega,\kappa)$
and $\varrho(\omega,\kappa)\asymp|\omega|^{1/2}+|\kappa|$. With
$\delta_\lambda(\omega,\kappa)=(\lambda^2\omega,\lambda\kappa)$, we define
\begin{equation}\label{eq:pre-lp-symbols}
    \chi_{-1}:=\chi,\qquad
    \chi_j:=\chi\circ\delta_{2^{-j-1}}
                  -\chi\circ\delta_{2^{-j}},\quad j\ge0.
\end{equation}
These symbols form a partition of unity in frequency. For $j\ge0$,
$\chi_j$ is supported where
$2^j\le\varrho\le\,2^{j+3}/3$, so only a fixed number of neighboring
blocks can overlap. Define
\begin{equation}\label{eq:pre-blocks}
\begin{split}
    \Delta_j^{\mathfrak s}f
       &:=\sum_{n,k}\chi_j(n,k)\widehat f(n,k)\varepsilon_{n,k},\\
    \Pi_J&:=\sum_{j=-1}^J\Delta_j^{\mathfrak s},
    \qquad
    m_J(n,k):=\varsigma(2^{-(J+1)}\varrho(n,k)).
\end{split}
\end{equation}
Thus $\Pi_J$ has the smooth symbol $m_J$,
$0\le m_J\le m_{J+1}\le1$, and $m_J\to1$ pointwise. The cutoff
retains temporal frequencies of order $2^{2J}$ and spatial frequencies
of order $2^J$. 

\begin{definition}[Parabolic H\"older--Besov spaces]\label{def:besov}
For $\beta\in\mathbb{R}$, let 
\[
    \mathcal{C}^\beta(\Lambda_T)
    :=
    B^{\beta,\mathfrak{s}}_{\infty,\infty}(\Lambda_T)
\]
be the inhomogeneous Besov space associated with the Littlewood--Paley 
decomposition $(\Delta_j^{\mathfrak{s}})_{j\ge -1}$, equipped with the norm
\[
    \|f\|_{\mathcal{C}^\beta(\Lambda_T)}
    :=
    \sup_{j\ge -1}
    2^{j\beta}
    \bigl\|\Delta_j^{\mathfrak{s}}f\bigr\|_{L^\infty(\Lambda_T)}.
\]
We denote by $\cC_g^\beta(\Lambda_T)$ the closure of
$C^\infty(\Lambda_T)$ in this norm. On $\R\times\T^d$, the same
decomposition, with continuous temporal frequency, defines
$\cC^\beta(\R\times\T^d)$. 
Let $\widetilde f\in\cC^\beta(\R\times\T^d)$ denote an extension of 
$f$ to $\R\times\T^d$. On the cylinder $ D_T$, we define
\[
    \|f\|_{\cC^\beta(\mathbb D_T)}
    :=
    \inf_{\substack{\widetilde f\in\cC^\beta(\R\times\T^d)\\
                    \widetilde f|_{ D_T}=f}}
    \|\widetilde f\|_{\cC^\beta(\R\times\T^d)}.
\]
For distributions, restriction is understood on the interior of the
cylinder; whenever the regularity is positive, the continuous
representative supplies the boundary values.
\end{definition}

Here, a periodic distribution $f$ on $\Lambda_T$
defines a distribution on $\R\times\T^d$, and restriction is bounded:
\[
    \|f|_{\DT}\|_{\cC^\beta(\DT)}
       \le C_{\beta,T}\|f\|_{\cC^\beta(\Lambda_T)}.
\]
For $\beta<0$, the space $\cC^\beta$ contains distributions
that need not be functions. For $0<\beta<1$, the anisotropic Besov space 
$B^{\beta,\mathfrak s}_{\infty,\infty}$ coincides with the anisotropic 
Hölder space $C^\beta_{\mathfrak s}$; see
\cite{Bownik2003,Triebel1992}. For $\beta\ge1$, the space is 
an anisotropic Zygmund-type space and is characterized by higher-order 
anisotropic differences, rather than by a single $d_{\mathfrak s}$-H\"older 
condition. We will only use the embedding
\[
    \cC^\beta(\DT)\hookrightarrow C^\gamma_{\mathfrak s}(\DT),
       \qquad 0<\gamma<\min\{1,\beta\}.
\]
Moreover, $L^\infty(\mathbb D_T)\hookrightarrow\cC^{-\kappa}(\mathbb D_T)$ 
for every $\kappa>0$: extending $f$ by zero outside the time interval and 
using the uniform $L^1$ bounds of the dyadic kernels, one obtains
\[
    \|\Delta_j^{\mathfrak s}\widetilde f\|_{L^\infty}
    \le
    C\|f\|_{L^\infty( D_T)},
\]
and hence
\[
    \|\widetilde f\|_{\cC^{-\kappa}(\R\times\T^d)}
    \le
    C\|f\|_{L^\infty( D_T)}.
\]
These properties and the equivalence of the resulting spaces under 
changes of smooth dyadic decompositions, are standard; see 
\cite{bahouri2011fourier,chandra2017spdes}.

For spatial regularity, we use the same construction with
$\chi^x(k)=\varsigma(|k|)$ and denote its blocks by $\Delta_j^x$.
Thus
\[
    \cC_x^s:=B^s_{\infty,\infty}(\T^d),\qquad
    \|f\|_{\cC_x^s}:=\sup_{j\ge-1}2^{js}\|\Delta_j^xf\|_\infty,
    \qquad
    \cC_{g,x}^s:=\overline{C^\infty(\T^d)}^{\,\cC_x^s}.
\]
The spatial cutoff $\Pi^x_{J_0}$ has symbol
$m^x_{J_0}(k)=\varsigma(2^{-(J_0+1)}|k|)$ and acts componentwise on
vector fields. We also define the Leray projector used in
Section~\ref{sec:nse}:
\begin{equation}\label{eq:pre-leray}
    \widehat{Pu}(k)
       :=\left(I-\frac{k\otimes k}{|k|^2}\right)\widehat u(k)
       \quad(k\ne0),\qquad \widehat{Pu}(0):=0.
\end{equation}
It projects onto divergence-free, mean-zero fields.

Throughout, $C,c>0$ may change from line to line. They depend on the
fixed equation, regularity exponents, and data bounds, but not on
truncation levels, approximation tolerances, or localization radii
unless stated otherwise. We write $A\lesssim B$ for $A\le CB$ and
$A\asymp B$ for the corresponding two-sided comparison.

\subsection{Gaussian forcing and tail estimates}

All random objects are defined on a complete probability space
$(\Omega,\mathcal F,\PP)$, with expectation $\E$.

\begin{assumption}[Gaussian forcing]\label{ass:noise}
There is a $\beta_0\in(-2,0)$ such that $\eta$ is a centered Gaussian
Borel random variable with values in
$ X_0:=\cC_g^{\beta_0}(\Lambda_T)$.
\end{assumption}
We use a weaker norm to measure perturbations of
the forcing. Fix
\begin{equation}\label{eq:exponents}
    -2<\beta_1<\beta_0<0,\qquad
    \delta:=\beta_0-\beta_1>0,\qquad
    X_1:=\cC^{\beta_1}(\DT),\qquad
    Y:=C(\DT).
\end{equation}
Restriction followed by the change of regularity gives a continuous
map $X_0\to X_1$. We suppress the restriction in the notation.
The norm on $Y$ is $\|u\|_Y=\sup_{(t,x)\in\DT}|u(t,x)|$.The strict gap $\delta$ will give the
Fourier truncation rate, while $\beta_1>-2$ ensures that the heat
convolution is continuous.

\begin{lemma}[Gaussian tails]\label{lem:fernique}
Under Assumption~\ref{ass:noise}, there are $c,C>0$ such that
\begin{equation}\label{eq:fernique}
    \E\exp\!\big(c\|\eta\|_{X_0}^2\big)<\infty,
    \qquad
    \PP(\|\eta\|_{X_0}>R)\le Ce^{-cR^2},
    \qquad R\ge0.
\end{equation}
In particular, $\E\|\eta\|_{X_0}^p<\infty$ for every $0<p<\infty$.
For every $q\ge0$, there are $c_q,C_q>0$, independent of $R$, such that
\begin{equation}\label{eq:weighted-tail}
    \E\left[(1+\|\eta\|_{X_0})^q
                \one_{\{\|\eta\|_{X_0}>R\}}\right]
       \le C_qe^{-c_qR^2},\qquad R\ge0.
\end{equation}
\end{lemma}

\begin{proof}
Write $Z=\|\eta\|_{X_0}$. Fernique's theorem gives an $a>0$ for which
$\E e^{aZ^2}<\infty$; see \cite[Chapter~2]{bogachev1998gaussian}.
By Markov's inequality applied to $e^{aZ^2}$, for every $R>0$,
\[
    \PP(Z>R)
    =
    \PP(e^{aZ^2}>e^{aR^2})
    \le
    e^{-aR^2}\E e^{aZ^2}.
\]
Moreover, since $|z|^p\le C_{p,a}e^{az^2}$ for every $p>0$ and some 
$C_{p,a}<\infty$, the finiteness of $\E e^{aZ^2}$ implies
\[
    \E|Z|^p
    \le
    C_{p,a}\E e^{aZ^2}
    <\infty.
\]
For the weighted tail, fix $q\ge0$ and choose $C_q$ such that
$(1+z)^q\le C_qe^{az^2/2}$ for $z\ge0$. On $\{Z>R\}$,
\[
    (1+Z)^q\le C_qe^{aZ^2}e^{-aR^2/2}.
\]
Taking expectations gives \eqref{eq:weighted-tail}, with
$c_q=a/2$ after adjusting $C_q$. In particular, these constants can
be chosen before the radius $R$.
\end{proof}

\subsection{The heat equation and the Schauder estimate}

Write $S(t)=e^{t\Delta}$. The periodic heat kernel is nonnegative
and has integral one, so
\[
    \|S(t)f\|_\infty\le\|f\|_\infty,\qquad t\ge0.
\]
For smooth forcing $\phi$, we define the heat convolution by
\begin{equation}\label{eq:pre-heat-ops}
    K\phi(t):=\int_0^t S(t-r)\phi(r)\,dr.
\end{equation}
For distributional forcing, the integral is interpreted through the
zero-initial-data heat equation, as justified in the next lemma. With
viscosity $\nu>0$, the notation $K_\nu$ means the same operator with
$S(t-r)$ replaced by $e^{\nu(t-r)\Delta}$.

\begin{lemma}[Schauder estimate]\label{lem:schauder}
Let $-2<\beta<0$. The heat equation
$(\partial_t-\Delta)v=\phi$, $v(0)=0$, defines a bounded linear map (solution map)
$K:\cC^\beta(\DT)\to\cC^{\beta+2}(\DT)$ with $v=K\phi$ satisfying
\begin{equation}\label{eq:schauder}
    \|K\phi\|_{\cC^{\beta+2}(\DT)}
       \le C_\beta\|\phi\|_{\cC^\beta(\DT)}.
\end{equation}
In particular, $K\phi$ is continuous and has zero initial trace. For
every $0<\gamma<\min\{1,\beta+2\}$,
\begin{equation}\label{eq:schauder-holder}
    \|K\phi\|_{C^\gamma_{\mathfrak s}(\DT)}
       \le C_{\beta,\gamma}\|\phi\|_{\cC^\beta(\DT)}.
\end{equation}
The same assertions hold for $K_\nu$, with constants also depending on
the viscosity coefficient $\nu>0$.
\end{lemma}

\begin{proof}
Set $\alpha=\beta+2\in(0,2)$. 
Choose an extension $\widetilde\phi\in\cC^\beta(\R\times\T^d)$ of 
$\phi$ such that
\[
    \|\widetilde\phi\|_{\cC^\beta(\R\times\T^d)}
    \le
    2\|\phi\|_{\cC^\beta( D_T)}.
\]
Let $\vartheta\in C_c^\infty(\R)$ satisfy 
$\vartheta\equiv1$ on a neighbourhood of $[0,T]$, and define
\[
    g(t,x)
    :=
    \vartheta(t)e^{-t}\widetilde\phi(t,x),
    \qquad
    (t,x)\in\R\times\T^d.
\]

Multiplication by this
fixed smooth function is bounded on $\cC^\beta$, so
\[
    \|g\|_{\cC^\beta(\R\times\T^d)}
       \le C_T\|\phi\|_{\cC^\beta(\DT)}.
\]
On the full cylinder $\mathbb{R}\times\mathbb{T}^d$, let
$w=(\partial_t-\Delta+1)^{-1}g$, defined by the Fourier multiplier
$(1+i\omega+|k|^2)^{-1}$.
For $j\ge0$, the parabolic block $\Delta_j^{\mathfrak s}$ localizes 
frequencies at scale $|\omega|\sim2^{2j}$ and $|k|\sim2^j$. On this 
region,
\[
    |1+i\omega+|k|^2|
    \gtrsim
    2^{2j},
\]
so the localized multiplier has convolution kernel with $L^1$-norm at 
most $C2^{-2j}$. Consequently,
\[
    \|\Delta_j^{\mathfrak s}w\|_{L^\infty}
    \le
    C2^{-2j}
    \sum_{\substack{\ell\ge-1\\|\ell-j|\le C}}
    \|\Delta_\ell^{\mathfrak s}g\|_{L^\infty}.
\] 
For the low-frequency block $j=-1$, the frequencies satisfy 
$|\omega|\lesssim1$ and $|k|\lesssim1$. The constant $1$ in the 
resolvent ensures that
\[
    |1+i\omega+|k|^2|
    \gtrsim
    1,
\]
so the multiplier $(1+i\omega+|k|^2)^{-1}$ is uniformly bounded and its 
localized kernel has uniformly bounded $L^1$-norm. Hence
\[
    \|\Delta_{-1}^{\mathfrak s}w\|_{L^\infty}
    \le
    C
    \sum_{\substack{\ell\ge-1\\|\ell+1|\le C}}
    \|\Delta_\ell^{\mathfrak s}g\|_{L^\infty}.
\]
By the dyadic characterization of $\mathcal C^\alpha$ and the identity
$\alpha-2=\beta$, for $j\ge0$ we have
\[
2^{j\alpha}\|\Delta_j^{\mathfrak s}w\|_{L^\infty}
\lesssim
\sum_{|\ell-j|\le N}
2^{(j-\ell)\beta}
\bigl(2^{\ell\beta}
\|\Delta_\ell^{\mathfrak s}g\|_{L^\infty}\bigr)
\lesssim \|g\|_{\mathcal C^\beta},
\]
since the sum contains only finitely many terms and $|j-\ell|\le C$.
Taking the supremum over $j\ge-1$ yields
\[
    \|w\|_{\cC^\alpha(\R\times\T^d)}
       \le C\|g\|_{\cC^\beta(\R\times\T^d)}.
\]
Since $\alpha>0$, $w$ has a continuous trace at $t=0$ and
\[
    \|w(0)\|_{\cC_x^\alpha}
       \le C\|w\|_{\cC^\alpha(\R\times\T^d)}.
\]
Recall the standard  estimate for heat extension (see \cite{bahouri2011fourier,chandra2017spdes,gubinelli2015paracontrolled})
\[
    \|S(\cdot)f\|_{\cC^\alpha(\DT)}
       \le C\|f\|_{\cC_x^\alpha},\qquad 0<\alpha<2.
\]
Note that for $\alpha=1$,
the spatial norm is the Zygmund norm, not the Lipschitz norm.

Now define
\[
    v(t):=e^t w(t)-S(t)w(0),\qquad 0\le t\le T.
\]
On $(0,T)\times\T^d$, it satisfies
$(\partial_t-\Delta)v=e^tg=\phi$ and $v(0)=0$. The preceding bounds
give $\|v\|_{\cC^\alpha(\DT)}
\le C\|\phi\|_{\cC^\beta(\DT)}$. Replacing $|k|^2$ by $\nu|k|^2$ in the
argument proves the result \eqref{eq:schauder-holder}. 
\end{proof}

\subsection{The semilinear solution map}

Let $F:\R\to\R$ be globally Lipschitz, with constant $L_F$. A mild
solution of \eqref{eq:she}, for deterministic forcing $\phi\in X_1=\cC^{\beta_1}(\DT)$,
is a function $u\in Y=C(\DT)$ such that
\begin{equation}\label{eq:mild}
    u(t)=S(t)u_0+K\phi(t)
               +\int_0^t S(t-r)F(u(r))\,dr.
\end{equation}

\begin{theorem}[Well-posedness and pathwise stability]\label{thm:stability}
For every $(u_0,\phi)\in C(\T^d)\times X_1$ there is a unique mild
solution $u=\cS(u_0,\phi)\in Y$. For two pairs of data,
\begin{equation}\label{eq:stability}
\begin{split}
    &\|\cS(u_0^1,\phi^1)-\cS(u_0^2,\phi^2)\|_Y\\
    &\qquad\le e^{L_FT}\left(
       \|u_0^1-u_0^2\|_\infty+
       C_{\beta_1}\|\phi^1-\phi^2\|_{X_1}\right).
\end{split}
\end{equation}
Moreover,
\begin{equation}\label{eq:apriori}
    \|\cS(u_0,\phi)\|_Y
       \le e^{L_FT}\left(
          \|u_0\|_\infty+
          C_{\beta_1}\|\phi\|_{X_1}+T|F(0)|
       \right).
\end{equation}
\end{theorem}

\begin{proof} By
lemma~\ref{lem:schauder}, we have $K\phi\in Y$ with zero initial trace.
Thus the right-hand side of \eqref{eq:mild} defines a map
$\mathcal S:Y\to Y$. Equip $Y$ with the equivalent norm
\[
    \|u\|_a:=\sup_{0\le t\le T}e^{-at}\|u(t)\|_\infty,
       \qquad a:=2L_F+1.
\]
Direct computation gives
\[
    e^{-at}\|\mathcal Su(t)-\mathcal Sv(t)\|_\infty
       \le L_F\int_0^t e^{-a(t-r)}\,dr\,\|u-v\|_a
       \le\frac{L_F}{a}\|u-v\|_a.
\]
Since $L_F/a<1/2$, the fixed-point theorem gives a unique solution
on the whole interval $[0,T]$.

Now for $u^i=\cS(u_0^i,\phi^i)$, subtracting the mild equations yields
\[
    \|u^1(t)-u^2(t)\|_\infty
       \le A+L_F\int_0^t\|u^1(r)-u^2(r)\|_\infty\,dr,
\]
where
$A=\|u_0^1-u_0^2\|_\infty+
C_{\beta_1}\|\phi^1-\phi^2\|_{X_1}$.
Gronwall's inequality gives
$\|u^1(t)-u^2(t)\|_\infty\le Ae^{L_Ft}$, which proves
\eqref{eq:stability} after taking the supremum over $t$.
Finally, the estimate $|F(r)|\le |F(0)|+L_F|r|$ gives
\[
    \|u(t)\|_\infty
       \le\|u_0\|_\infty+C_{\beta_1}\|\phi\|_{X_1}
             +T|F(0)|+L_F\int_0^t\|u(r)\|_\infty\,dr.
\]
Then \eqref{eq:apriori} follows from the Gronwall inequality. 
\end{proof}

\subsection{Uniform bounds on bounded sets of data}

\begin{assumption}[Initial-data class]\label{ass:initial}
For fixed $M,L_0>0$, let
\[
    \cG_0:=\left\{u_0\in C(\T^d):
       \|u_0\|_\infty\le M,\quad
       |u_0(x)-u_0(y)|\le L_0d_{\T^d}(x,y)
                      \text{ for all }x,y\in\T^d\right\}.
\]
\end{assumption}

For the approximation of the output functions, we also need a
modulus of continuity that is uniform over bounded forcing sets.

\begin{lemma}[Uniform solution bounds]\label{lem:modulus}
Fix
\begin{equation}\label{eq:gamma}
    0<\gamma<\min\{1,\beta_0+2\}.
\end{equation}
There are increasing affine functions
$B,\Lambda:[0,\infty)\to[1,\infty)$, depending only on
$d$, $T$, $\beta_0$, $\beta_1$, $\gamma$, $M$, $L_0$, $L_F$, and
$|F(0)|$, such that for
every $R\ge0$, $u_0\in\cG_0$, and $\zeta\in X_0$ with
$\|\zeta\|_{X_0}\le R$,
\begin{equation}\label{eq:modulus}
\begin{split}
    \|\cS(u_0,\zeta)\|_Y&\le B(R),\\
    |\cS(u_0,\zeta)(z)-\cS(u_0,\zeta)(z')|
       &\le\Lambda(R)d_{\mathfrak s}(z,z')^\gamma.
\end{split}
\end{equation}
In particular, $B(R)+\Lambda(R)\le C(1+R)$.
\end{lemma}

\begin{proof}
Let $C_{\rm res}$ be a bound for restriction from $X_0$ to $X_1$.
The a priori estimate \eqref{eq:apriori} gives the first assertion with
\[
    B(R):=1+e^{L_FT}
       \big(M+C_{\beta_1}C_{\rm res}R+T|F(0)|\big).
\]
Let \(L_0=\operatorname{Lip}(u_0)\). Then $\operatorname{Lip}(S(t)u_0)\le L_0$
\[
\begin{aligned}
\|S(h)u_0-u_0\|_\infty
&\le L_0\int_{\mathbb T^d}
K_h(z)\,d_{\mathbb T^d}(z,0)\,dz\\
&\le C L_0 h^{1/2}.
\end{aligned}
\]
For \(t\ge s\ge0\),
\[
\begin{aligned}
|S(t)u_0(x)-S(s)u_0(y)|
&\le |S(t)u_0(x)-S(t)u_0(y)|
   +\|S(s)(S(t-s)u_0-u_0)\|_\infty\\
&\le L_0 d_{\mathbb T^d}(x,y)
   +\|S(t-s)u_0-u_0\|_\infty\\
&\le C L_0\bigl(d_{\mathbb T^d}(x,y)+|t-s|^{1/2}\bigr).
\end{aligned}
\]

Writing \(D=\operatorname{diam}_{\mathfrak s}(\DT)<\infty\), we obtain,
for every \(0<\gamma<1\),
\[
[S(\cdot)u_0]_{C^\gamma_{\mathfrak s}(\DT)}
\le D^{1-\gamma}
   [S(\cdot)u_0]_{C^1_{\mathfrak s}(\DT)}
\le C L_0,
\qquad
\|S(\cdot)u_0\|_\infty\le M.
\]
Hence
\[
\|S(\cdot)u_0\|_{C^\gamma_{\mathfrak s}(\DT)}
\le C(M+L_0),
\]
uniformly up to \(t=0\).

By Lemma~\ref{lem:schauder} with \(\beta=\beta_0\) and
\(\gamma<\beta_0+2\), as in \eqref{eq:gamma},
\[
\|K\zeta\|_{C^\gamma_{\mathfrak s}(\DT)}
\le C\|\zeta\|_{\cC^{\beta_0}(\DT)}
\le CR.
\]
Moreover,
\[
\|F(u)\|_\infty
\le |F(0)|+L_F\|u\|_\infty
\le |F(0)|+L_FB(R).
\]
Since
\[
L^\infty\hookrightarrow\cC^{-1/2},
\qquad
\gamma<1<-\tfrac12+2=\tfrac32,
\]
another application of Lemma~\ref{lem:schauder} gives
\[
\begin{aligned}
\|K(F(u))\|_{C^\gamma_{\mathfrak s}(\DT)}
&\le C\|F(u)\|_{\cC^{-1/2}(\DT)}\\
&\le C\|F(u)\|_\infty\\
&\le C\bigl(|F(0)|+L_FB(R)\bigr).
\end{aligned}
\]
Combining these estimates proves the second assertion with
\[
\Lambda(R)
:=1+C\bigl(M+L_0+R+|F(0)|+L_FB(R)\bigr).
\]
Both \(B(R)\) and \(\Lambda(R)\) are affine in \(R\), with
coefficients depending only on the fixed data in the statement.
\end{proof}

\section{Finite-dimensional encoding and operator architecture}
\label{sec:reduction}

\subsection{A real finite-rank measurement map}

Let $\Pi_J=\sum_{j\le J}\Delta_j^{\mathfrak s}$ and let $m_J$ denote its
smooth, even Fourier multiplier.  We choose the Littlewood--Paley
decomposition with nested low-pass multipliers, so that
\begin{equation}\label{eq:nested-cutoffs}
    0\le m_J(\lambda)\le m_{J+1}(\lambda)\le1,
    \qquad m_J(\lambda)\longrightarrow1
    \quad\text{for every fixed }\lambda.
\end{equation}
The operator \(\Pi_J\) is a finite-rank smoothing cutoff,
not necessarily a projection.  Let \((e_\lambda)_\lambda\) be the real orthonormal trigonometric
basis of \(L^2(\Lambda_T)\), consisting of the normalized constant
function and the normalized sine and cosine functions, with one
sine--cosine pair for each pair of nonzero frequencies
\(\{\theta,-\theta\}\).
Since \(m_J(\theta)=m_J(-\theta)\), the operator \(\Pi_J\)
multiplies both functions in each pair by the same scalar
\(m_J(\theta)\). Thus every \(e_\lambda\) is an eigenfunction
of \(\Pi_J\).

Let \(\Lambda_J\) be the set of indices of the basis functions
not annihilated by \(\Pi_J\):
\[
\Lambda_J:=\{\lambda:\Pi_J e_\lambda\neq0\}.
\]  Define
\begin{equation}\label{eq:sensors}
    c_J(\zeta):=(\langle\zeta,e_\lambda\rangle)_{\lambda\in\Lambda_J}
       \in\R^{N_J},
    \qquad
    \Rec_J(c):=\sum_{\lambda\in\Lambda_J}
       m_J(\lambda)c_\lambda e_\lambda.
\end{equation}
Then $\Rec_J(c_J(\zeta))=\Pi_J\zeta$ exactly.  Parabolic frequency counting
gives
\begin{equation}\label{eq:rank}
    N_J:=|\Lambda_J|\asymp 2^{J(d+2)}.
\end{equation}
The count and the uniform boundedness of the smooth cutoffs will be proved in
Lemma~\ref{lem:parabolic-count}.
For an abstract forcing satisfying Assumption~\ref{ass:noise}, $c_J(\eta)$ is
a centered jointly Gaussian vector with covariance
$\Sigma_{\lambda\mu}=\E[\langle\eta,e_\lambda\rangle
\langle\eta,e_\mu\rangle]$.  No independence of these coordinates is assumed
in the approximation theorems.
\begin{lemma}\label{lem:parabolic-count}
For \(J\ge0\),
\begin{equation}\label{eq:appendix-rank}
    \operatorname{rank}(\Pi_J)\asymp 2^{J(d+2)}.
\end{equation}
Moreover, for every \(\beta\in\R\), there exists \(C_\beta>0\),
independent of \(J\), such that
\begin{equation}\label{eq:appendix-cutoff-bound}
    \|\Pi_J f\|_{\cC^\beta}
    \le C_\beta\|f\|_{\cC^\beta}.
\end{equation}
\end{lemma}

\begin{proof}
For \(a>0\), set
\[
    Q_J(a):=
    \bigl\{(n,k)\in\Z\times\Z^d:
    |n|\le a2^{2J},\ |k|\le a2^J\bigr\}.
\]
The parabolic cutoff construction gives fixed constants \(0<c<C\)
such that
\[
    Q_J(c)
    \subseteq \{(n,k):m_J(n,k)\neq0\}
    \subseteq Q_J(C).
\]
Since
\[
    |Q_J(a)|\asymp 2^{2J}2^{dJ}=2^{J(d+2)},
\]
and the rank of \(\Pi_J\) equals the number of frequencies on
which \(m_J\) is nonzero, \eqref{eq:appendix-rank} follows.

For the uniform bound, finite overlap of the dyadic supports yields
a fixed integer \(N\) such that
\[
    \Delta_j^{\mathfrak s}\Pi_J f
    =
    \sum_{|j-j'|\le N}
    \Delta_j^{\mathfrak s}\Pi_J
    \Delta_{j'}^{\mathfrak s}f.
\]
The operators \(\Delta_j^{\mathfrak s}\Pi_J\) have convolution
kernels with \(L^1\) norms bounded uniformly in \(j\) and \(J\).
Consequently,
\[
\begin{aligned}
    2^{j\beta}
    \|\Delta_j^{\mathfrak s}\Pi_J f\|_\infty
    &\le
    C\sum_{|j-j'|\le N}
    2^{(j-j')\beta}
    2^{j'\beta}
    \|\Delta_{j'}^{\mathfrak s}f\|_\infty\\
    &\le C_\beta\|f\|_{\cC^\beta}.
\end{aligned}
\]
Taking the supremum over \(j\) proves
\eqref{eq:appendix-cutoff-bound}.
\end{proof}

\begin{lemma}\label{lem:projection}
For every $\zeta\in\cC^{\beta_0}(\Lambda_T)$,
\begin{equation}\label{eq:projection}
    \|(I-\Pi_J)\zeta\|_{\cC^{\beta_1}}
    \le C2^{-J\delta}\|\zeta\|_{\cC^{\beta_0}}.
\end{equation}
\end{lemma}

\begin{proof}
Almost orthogonality gives
$\Delta_j^{\mathfrak s}(I-\Pi_J)\zeta=0$ for $j<J-N$, then
\[
2^{j\beta_1}\|\Delta_j^{\mathfrak s}(I-\Pi_J)\zeta\|_\infty
\le C2^{-j(\beta_0-\beta_1)}
   2^{j\beta_0}\|\Delta_j^{\mathfrak s}\zeta\|_\infty
\]
for $j\ge J-N$.  Taking the supremum over $j$ proves the claim.
\end{proof}

\begin{lemma}\label{lem:coeff}
There is a constant \(C>0\), independent of \(J\), such that
\begin{align}
    \|c_J(\zeta)\|_{\ell^\infty}
    &\le C2^{-J\beta_0}\|\zeta\|_{\cC^{\beta_0}},
    \label{eq:coeff-bound}\\
    \|\Rec_J(c)-\Rec_J(c')\|_{X_1}
    &\le C N_J^{1/2}\|c-c'\|_{\ell^2}.
    \label{eq:rec-bound}
\end{align}
\end{lemma}

\begin{proof}
Let \(e_\lambda\) have parabolic frequency comparable to \(2^j\).
By finite overlap of the Littlewood--Paley supports,
\[
    |\langle\zeta,e_\lambda\rangle|
    \le C\sum_{|j'-j|\le N}
    \|\Delta_{j'}^{\mathfrak s}\zeta\|_\infty
    \le C2^{-j\beta_0}\|\zeta\|_{\cC^{\beta_0}},
\]
where \(N\) is fixed; the constant mode satisfies the corresponding
low-frequency bound.
For \(\lambda\in\Lambda_J\), we have \(j\le J+C_0\).
Since \(\beta_0<0\),
\[
    \sup_{\lambda\in\Lambda_J}
    |\langle\zeta,e_\lambda\rangle|
    \le C2^{-J\beta_0}\|\zeta\|_{\cC^{\beta_0}},
\]
which proves \eqref{eq:coeff-bound}.

Since \(\beta_1<0\), the normalized trigonometric basis functions satisfy
\[
    \sup_\lambda\|e_\lambda\|_{X_1}<\infty.
\]
Using \(0\le m_J\le1\), the triangle inequality and
Cauchy--Schwarz give
\[
\begin{aligned}
    \|\Rec_J(c)-\Rec_J(c')\|_{X_1}
    &\le
    \sum_{\lambda\in\Lambda_J}
    |m_J(\lambda)|\,|c_\lambda-c'_\lambda|
    \|e_\lambda\|_{X_1}\\
    &\le C\sum_{\lambda\in\Lambda_J}|c_\lambda-c'_\lambda|\\
    &\le C N_J^{1/2}\|c-c'\|_{\ell^2}.
\end{aligned}
\]
This proves \eqref{eq:rec-bound}.
\end{proof}

\subsection{Initial-data measurements and the reduced map}

Let $x_1,\ldots,x_{N_0}$ be a periodic tensor grid of mesh $\delta_0$ and let
$\vec u_0=(u_0(x_i))_{i\le N_0}$.  Denote by $\Rec_0(\vec v)$ the periodic
piecewise multilinear reconstruction.  
\begin{lemma}
\label{lem:initial-interpolation}
Let $(x_i)_{i\le N_0}$ be a periodic tensor grid on $\T^d$ with mesh at most
$\delta_0$, and let $\Rec_0$ be its piecewise multilinear interpolant.  If
$u_0$ is $L_0$-Lipschitz, then
\begin{equation}\label{eq:appendix-interpolation}
    \|u_0-\Rec_0((u_0(x_i))_i)\|_\infty
    \le C_dL_0\delta_0.
\end{equation}
For arbitrary nodal vectors $v,v'$, one has
\begin{equation}\label{eq:appendix-interpolation-stability}
    \|\Rec_0(v)-\Rec_0(v')\|_\infty
    \le|v-v'|_{\ell^\infty}.
\end{equation}
Moreover, $N_0\asymp\delta_0^{-d}$.
\end{lemma}

\begin{proof}
On each grid cell, multilinear interpolation has the form
\[
    \Rec_0((u_0(x_i))_i)(x)
    =\sum_{\nu\in\{0,1\}^d}w_\nu(x)u_0(x_\nu),
\]
where the vertex weights are nonnegative and sum to one.  Every vertex
$x_\nu$ of the cell satisfies
$d_{\T^d}(x,x_\nu)\le\sqrt d\,\delta_0$.  Therefore
\[
\left|u_0(x)-\Rec_0((u_0(x_i))_i)(x)\right|
\le\sum_\nu w_\nu(x)L_0d_{\T^d}(x,x_\nu)
\le\sqrt d\,L_0\delta_0.
\]
The same convex-combination formula applied to $v-v'$ proves
\eqref{eq:appendix-interpolation-stability}.  Finally, a periodic tensor grid
has a number of nodes comparable to the reciprocal of its cell volume,
namely $\delta_0^{-d}$.
\end{proof}

For $z=(\vec v,c)\in\R^{N_0}\times\R^{N_J}$ define
\begin{equation}\label{eq:GammaJ}
    \Gamma_J(z)
    :=\cS\big(\Rec_0(\vec v),\Rec_J(c)\big)\in Y.
\end{equation}

\begin{lemma}\label{lem:Gamma-lip}
The map \(\Gamma_J\) is Lipschitz, with
\begin{equation}\label{eq:Gamma-lip}
    \|\Gamma_J(\vec v,c)-\Gamma_J(\vec v',c')\|_Y
    \le L_J\bigl(
    \|\vec v-\vec v'\|_{\ell^\infty}
    +\|c-c'\|_{\ell^2}\bigr),
\end{equation}
where
\[
    L_J:=Ce^{L_FT}\max\{1,N_J^{1/2}\},
\]
and \(C\) is independent of \(J\).
\end{lemma}

\begin{proof}
By Theorem~\ref{thm:stability}, \eqref{eq:appendix-interpolation-stability}, and
\eqref{eq:rec-bound},
\[
\begin{aligned}
    \|\Gamma_J(\vec v,c)-\Gamma_J(\vec v',c')\|_Y
    &\le Ce^{L_FT}\bigl(
    \|\vec v-\vec v'\|_{\ell^\infty}
    +N_J^{1/2}\|c-c'\|_{\ell^2}\bigr)\\
    &\le L_J\bigl(
    \|\vec v-\vec v'\|_{\ell^\infty}
    +\|c-c'\|_{\ell^2}\bigr).
\end{aligned}
\]
\end{proof}

\begin{proposition}
\label{prop:reduction-error}
There is a constant $C$, independent of $J$ and $\delta_0$, such that for
every $u_0\in\cG_0$ and $\zeta\in X_0$,
\begin{equation}\label{eq:reduction-error}
\left\|
\cS(u_0,\zeta)
-\Gamma_J\big(\vec u_0,c_J(\zeta)\big)
\right\|_Y \le C e^{L_FT}
\left(L_0\delta_0
      +2^{-J\delta}\|\zeta\|_{X_0}\right).
\end{equation}
Equivalently, in terms of the number of forcing measurements,
\begin{equation}\label{eq:reduction-rank-rate}
\left\|
\cS(u_0,\zeta)
-\Gamma_J\big(\vec u_0,c_J(\zeta)\big)
\right\|_Y
\le C e^{L_FT}
\left(L_0\delta_0
      +N_J^{-\delta/(d+2)}\|\zeta\|_{X_0}\right).
\end{equation}
\end{proposition}

\begin{proof}
By \eqref{eq:sensors} and \eqref{eq:GammaJ},
\[
    \Gamma_J\big(\vec u_0,c_J(\zeta)\big)
    =\cS\big(\Rec_0(\vec u_0),\Pi_J\zeta\big).
\]
The stability estimate \eqref{eq:stability} therefore bounds the left-hand
side of \eqref{eq:reduction-error} by a fixed multiple of
\[
    \|u_0-\Rec_0(\vec u_0)\|_\infty
    +\|(I-\Pi_J)\zeta\|_{X_1}.
\]
Apply \eqref{eq:appendix-interpolation} to the first term and
Lemma~\ref{lem:projection} to the second.  This proves
\eqref{eq:reduction-error}.  Finally,
$N_J\asymp2^{J(d+2)}$ implies
$2^{-J\delta}\asymp N_J^{-\delta/(d+2)}$, which yields
\eqref{eq:reduction-rank-rate}.
\end{proof}

\begin{remark}
The exponent $\delta=\beta_0-\beta_1$ is the amount of regularity exchanged
for compactness.  A larger gap improves the decay in
\eqref{eq:reduction-rank-rate}, but the stability space must still satisfy
$\beta_1>-2$ so that the heat convolution is continuous.  Hence
$\delta<\beta_0+2$.  The limiting sensor exponent permitted by this argument
is therefore $(\beta_0+2)/(d+2)$, approached by taking
$\beta_1\downarrow-2$.
\end{remark}

\subsection{Deep neural operator architecture}

Let $\sigma(r)=\max\{r,0\}$ denote the ReLU activation, applied
coordinatewise to vectors.

\begin{definition}[ReLU network class]\label{def:FNN}
Let $D\subset\R^n$ be compact.  We write
\[
    \Phi\in\FNN(n,m,L,p,K,\kappa,R;D)
\]
if $\Phi:\R^n\to\R^m$ has a representation
\begin{align*}
    x_0&=x,\\
    x_\ell&=\sigma(W_\ell x_{\ell-1}+b_\ell),
        \qquad 1\le\ell<L,\\
    \Phi(x)&=W_Lx_{L-1}+b_L,
\end{align*}
such that every hidden layer has width at most $p$, the matrices and biases
have at most $K$ nonzero entries in total, every parameter has absolute value
at most $\kappa$, and $\|\Phi\|_{L^\infty(D)}\le R$.  Thus $L$, $p$, $K$,
and $\kappa$ record depth, width, sparsity, and parameter magnitude,
respectively.
\end{definition}

This is the network model used in \cite{bayraktar2025deep}.  We shall use the
following elementary constructions.

\begin{lemma}\label{lem:relu-primitives}
The following statements hold.
\begin{enumerate}
    \item For $a<b$, the projection onto $[a,b]$ is represented exactly by
    \[
        P_{[a,b]}(x)=a+\sigma(x-a)-\sigma(x-b).
    \]
    Applying this formula coordinatewise gives an exact ReLU realization of
    the projection onto a rectangular box.
    \item For $c\in\R$ and $r>0$, the hat function
    \[
        H_{c,r}(x)
        =\sigma\!\left(1-r^{-1}
          \big(\sigma(x-c)+\sigma(c-x)\big)\right)
        =\left(1-\frac{|x-c|}{r}\right)_+
    \]
    is represented exactly by a fixed-depth ReLU network.
    \item Let $M\ge1$ and $\tau\in(0,1)$.  There is a ReLU network
    $\operatorname{Mult}_{M,\tau}:\R^2\to\R$ such that
    \[
        \sup_{|x|,|y|\le M}
        |\operatorname{Mult}_{M,\tau}(x,y)-xy|\le\tau.
    \]
    Its width is bounded by an absolute constant, and its depth and number of
    nonzero parameters are $O_M(\log(\tau^{-1}))$.
\end{enumerate}
Consequently, for every integer $q\ge2$, a product of $q$ factors in $[0,1]$
can be approximated within
$\tau$ by a ReLU network with depth
$O(\log q\,\log(q/\tau))$ and $O(q\log(q/\tau))$ nonzero parameters.
For fixed $q$, both quantities are $O_q(\log(1/\tau))$.
\end{lemma}

\begin{proof}
The first two statements follow from
\[
    \sigma(x)=\max\{x,0\},
    \qquad
    \sigma(x-c)+\sigma(c-x)=|x-c|.
\]
The third statement is \cite[Lemma~3.1]{bayraktar2025deep}.

For the product approximation, let
\[
    Q:=2^{\lceil\log_2 q\rceil},
    \qquad q\le Q<2q,
\]
and pad the \(q\) inputs with \(Q-q\) factors equal to \(1\).
Arrange the \(Q\) factors as the leaves of a balanced binary tree.
At each internal node, use
\[
    (u,v)\longmapsto
    P_{[0,1]}\bigl(\operatorname{Mult}_{1,\delta}(u,v)\bigr),
    \qquad
    \delta:=\frac{\tau}{2q}.
\]

To estimate the error, let \(a,b\in[0,1]\) be the exact products
associated with the two child nodes, and let
\(\widetilde a,\widetilde b\in[0,1]\) be their approximations.
Since \(P_{[0,1]}\) is \(1\)-Lipschitz and fixes \(ab\),
\[
\begin{aligned}
    &\left|
    P_{[0,1]}\bigl(
    \operatorname{Mult}_{1,\delta}(\widetilde a,\widetilde b)
    \bigr)-ab
    \right|\\
    &\qquad\le
    \left|
    \operatorname{Mult}_{1,\delta}(\widetilde a,\widetilde b)
    -\widetilde a\widetilde b
    \right|
    +|\widetilde a\widetilde b-ab|\\
    &\qquad\le
    \delta+|\widetilde a-a|+|\widetilde b-b|.
\end{aligned}
\]
The leaf errors vanish, so induction over the tree gives a root error
of at most
\[
    (Q-1)\delta<2q\delta=\tau.
\]

The tree has \(\log_2 Q=O(\log q)\) levels and \(Q-1=O(q)\)
internal nodes. Each node requires depth and number of nonzero
parameters \(O(\log(q/\tau))\), including the exact projection onto $[0,1]$.
Implementing the nodes in parallel at each level therefore gives
\[
    \operatorname{depth}
    =O\bigl(\log q\,\log(q/\tau)\bigr),
    \qquad
    \operatorname{size}
    =O\bigl(q\log(q/\tau)\bigr).
\]
For fixed \(q\), both bounds are \(O_q(\log(1/\tau))\).
\end{proof}

For fixed $J$ and $\delta_0$, define the measurement map
\begin{equation}\label{eq:encoder}
    \mathcal E_{J,\delta_0}(u_0,\zeta)
    :=(\vec u_0,c_J(\zeta))\in\R^{N_0+N_J}.
\end{equation}
The branch--trunk architecture considered below has the form
\begin{equation}\label{eq:architecture}
    \Gamma_\theta\big(\mathcal E_{J,\delta_0}(u_0,\zeta)\big)(y)
    =\sum_{\ell=1}^{N_1}
      \widetilde a_\ell\big(\vec u_0,c_J(\zeta);\theta_b\big)
      \widetilde q_\ell(y;\theta_t),
    \qquad y=(t,x)\in\DT.
\end{equation}
The branch networks $\widetilde a_\ell$ act on the finite measurement vector,
whereas the trunk networks $\widetilde q_\ell$ act on the output coordinate.

\subsection{A finite-dimensional partition}

The partition of unity is applied only after the measurement map has placed the input
in a finite-dimensional Euclidean space.

\begin{lemma}\label{lem:pou}
Let $K\subset\R^N$ be compact, let $\Gamma:K\to Y$ be $L$-Lipschitz with
respect to the Euclidean metric, and let $z^1,\ldots,z^m$ be an $r/2$-net of
$K$.  Define
\[
    h_k(z):=\left(1-\frac{|z-z^k|_2}{r}\right)_+,
    \qquad
    \rho_k(z):=\frac{h_k(z)}{\sum_{j=1}^m h_j(z)}.
\]
Then the denominator is at least $1/2$ on $K$, the functions $\rho_k$ form a
continuous partition of unity, and
\begin{equation}\label{eq:pou-error}
    A_r(z):=\sum_{k=1}^m\rho_k(z)\Gamma(z^k)
    \quad\Longrightarrow\quad
    \sup_{z\in K}\|\Gamma(z)-A_r(z)\|_Y\le Lr.
\end{equation}
\end{lemma}

\begin{proof}
Every $z\in K$ lies within $r/2$ of a center, so the denominator is at least
$1/2$.  If $\rho_k(z)>0$, then $|z-z^k|_2<r$.  The estimate follows from
convexity:
\[
    \|\Gamma(z)-A_r(z)\|_Y
    \le\sum_k\rho_k(z)\|\Gamma(z)-\Gamma(z^k)\|_Y
    \le Lr.
\]
\end{proof}

\subsection{Relation to Wiener chaos expansions}

For the linear equation, the Fourier measurements determine a finite sum in the first Wiener chaos.  The following result also gives an exact covariance formula for the mean-square truncation error.

\begin{lemma}
\label{lem:wce-linear}
Assume $F\equiv0$.  For every real Fourier mode $e_\lambda$, set
\[
    \xi_\lambda:=\langle\eta,e_\lambda\rangle,
    \qquad z_\lambda:=Ke_\lambda,
\]
and write
$\Sigma_{\lambda\mu}:=\E[\xi_\lambda\xi_\mu]$.  Then:
\begin{enumerate}
    \item For every $J$,
    \begin{equation}\label{eq:wce-finite}
        \cS(u_0,\Pi_J\eta)
        =e^{t\Delta}u_0
         +\sum_{\lambda\in\Lambda_J}
          m_J(\lambda)\xi_\lambda z_\lambda.
    \end{equation}
    \item The solutions in \eqref{eq:wce-finite} converge to
    $\cS(u_0,\eta)$ almost surely and in
    $L^2(\Omega;C(\DT))$.  The random part $K\eta$ belongs to the first
    Wiener chaos generated by $(\xi_\lambda)_\lambda$; all chaos components
    of order at least two vanish.
    \item The truncation error has the exact covariance representation
    \begin{align}
    &\E\|\cS(u_0,\eta)-\cS(u_0,\Pi_J\eta)\|_{L^2(\DT)}^2
        \label{eq:wce-covariance}\\
    &\quad=
      \lim_{L\to\infty}\;\sum_{\lambda,\mu\in\Lambda_L}
      (m_L(\lambda)-m_J(\lambda))(m_L(\mu)-m_J(\mu))
      \Sigma_{\lambda\mu}
      \langle z_\lambda,z_\mu\rangle_{L^2(\DT)},
    \end{align}
    each inner sum being finite and exact for the smooth outer cutoff
    $\Pi_L$, $L>J$.  If the real
    Fourier basis diagonalizes the covariance, so that
    $\Sigma_{\lambda\mu}=\sigma_\lambda^2\one_{\{\lambda=\mu\}}$, then
    \begin{equation}\label{eq:wce-diagonal}
      \E\|\cS(u_0,\eta)-\cS(u_0,\Pi_J\eta)\|_{L^2(\DT)}^2
      =\sum_\lambda(1-m_J(\lambda))^2
        \sigma_\lambda^2\|z_\lambda\|_{L^2(\DT)}^2.
    \end{equation}
\end{enumerate}
\end{lemma}

\begin{proof}
For \(F\equiv0\),
\[
    \cS(u_0,\phi)=e^{t\Delta}u_0+K\phi.
\]
Since
\[
    \Pi_J\eta
    =\sum_{\lambda\in\Lambda_J}
      m_J(\lambda)\xi_\lambda e_\lambda,
\]
linearity of \(K\) gives
\[
    \cS(u_0,\Pi_J\eta)
    =e^{t\Delta}u_0+
      \sum_{\lambda\in\Lambda_J}
      m_J(\lambda)\xi_\lambda z_\lambda,
\]
proving \eqref{eq:wce-finite}.

Theorem~\ref{thm:stability} with \(L_F=0\) and
Lemma~\ref{lem:projection} yield
\begin{equation}\label{eq:wce-convergence}
    \|\cS(u_0,\eta)-\cS(u_0,\Pi_J\eta)\|_Y
    \le C2^{-J\delta}\|\eta\|_{X_0}.
\end{equation}
By Lemma~\ref{lem:fernique},
\[
    \|\eta\|_{X_0}<\infty
    \quad\text{almost surely},
    \qquad
    \E\|\eta\|_{X_0}^2<\infty.
\]
Thus \eqref{eq:wce-convergence} implies convergence both almost
surely and in \(L^2(\Omega;C(\DT))\).

Each random part
\[
    K\Pi_J\eta
    =\sum_{\lambda\in\Lambda_J}
      m_J(\lambda)\xi_\lambda z_\lambda
\]
is a finite linear combination of the centered Gaussian coordinates
\(\xi_\lambda\), with deterministic coefficients \(z_\lambda\).
Since the first Wiener chaos is closed in \(L^2\), the limit
\(K\eta\) belongs to the first chaos generated by these coordinates.
Consequently, the solution has only its deterministic component
\(e^{t\Delta}u_0\) and its first-chaos component \(K\eta\).

Next fix \(J\) and let \(L>J\). By nesting,
\(\Lambda_J\subseteq\Lambda_L\), so
\[
    K(\Pi_L-\Pi_J)\eta
    =
    \sum_{\lambda\in\Lambda_L}
    \bigl(m_L(\lambda)-m_J(\lambda)\bigr)
    \xi_\lambda z_\lambda.
\]
Expanding the squared norm and taking expectations gives
\[
\begin{aligned}
    &\E\|K(\Pi_L-\Pi_J)\eta\|_{L^2(\DT)}^2\\
    &\quad=
    \sum_{\lambda,\mu\in\Lambda_L}
    \bigl(m_L(\lambda)-m_J(\lambda)\bigr)
    \bigl(m_L(\mu)-m_J(\mu)\bigr)
    \Sigma_{\lambda\mu}
    \langle z_\lambda,z_\mu\rangle_{L^2(\DT)}.
\end{aligned}
\]
Moreover, \eqref{eq:wce-convergence} implies
\[
    K(\Pi_L-\Pi_J)\eta
    \longrightarrow K(I-\Pi_J)\eta
    \quad\text{in }L^2(\Omega;L^2(\DT)).
\]
The second moments therefore converge, proving
\eqref{eq:wce-covariance}.

If \(\Sigma\) is diagonal, the finite covariance identity reduces to
\[
    \E\|K(\Pi_L-\Pi_J)\eta\|_{L^2(\DT)}^2
    =
    \sum_{\lambda\in\Lambda_L}
    \bigl(m_L(\lambda)-m_J(\lambda)\bigr)^2
    \sigma_\lambda^2\|z_\lambda\|_{L^2(\DT)}^2.
\]
For each \(\lambda\), the nesting property
\eqref{eq:nested-cutoffs} gives
\[
    \bigl(m_L(\lambda)-m_J(\lambda)\bigr)^2
    \uparrow
    \bigl(1-m_J(\lambda)\bigr)^2
    \qquad\text{as }L\to\infty.
\]
Extending each finite sum by zero outside \(\Lambda_L\) and applying
monotone convergence proves \eqref{eq:wce-diagonal}.
\end{proof}

\section{Approximation results}\label{sec:approximation}

\subsection{Function approximation: the trunk network}\label{sec:trunk}

We first approximate functions $v(t,x)$ on
$\DT=[0,T]\times\T^d$ using their values at finitely many grid
points. The H\"older estimate below controls their variation
in terms of $|t-s|^{1/2}+d_{\T^d}(x,y)$. We therefore choose
a spatial mesh of size $h$ and a temporal mesh of size $h^2$,
so that both contributions are of order $h$ within each grid cell.

\begin{theorem}[Parabolic function approximation]\label{thm:trunk}
Let $B,\Lambda\ge1$, $0<\gamma\le1$, and let $\mathcal V$ be a family of
functions $v\in C(\DT)$ satisfying
\[
    \|v\|_\infty\le B,
    \qquad |v(z)-v(z')|\le\Lambda d_{\mathfrak s}(z,z')^\gamma.
\]
For every $\varepsilon_T\in(0,B]$ with $\varepsilon_T\le2\Lambda$, there are
parabolic grid nodes $z_1,\ldots,z_{N_1}\in\DT$ and ReLU networks
$\widetilde q_1,\ldots,\widetilde q_{N_1}:\R^{1+d}\to\R$ such that
\begin{align}
    \sup_{v\in\mathcal V}\sup_{z\in\DT}
    \left|v(z)-\sum_{\ell=1}^{N_1}v(z_\ell)\widetilde q_\ell(z)\right|
        &\le\varepsilon_T,\label{eq:trunk-error}\\
    \sup_{z\in\DT}\sum_{\ell=1}^{N_1}|\widetilde q_\ell(z)|
        &\le\frac32.\label{eq:trunk-l1}
\end{align}
Moreover,
\begin{equation}\label{eq:trunk-count}
    N_1\asymp
    \left(\frac{\Lambda}{\varepsilon_T}\right)^{(d+2)/\gamma}.
\end{equation}
Each trunk network may be chosen with depth and number of nonzero parameters
$O((d+1)\log(BN_1/\varepsilon_T))$, with constants depending on the fixed
domain and $d$.
\end{theorem}

\begin{proof}
We first construct an interpolant from finitely many point values,
then approximate its basis functions by ReLU networks.
Choose
\begin{equation}\label{eq:trunk-h}
    h:=\frac{1}{1+\sqrt d}
       \left(\frac{\varepsilon_T}{2\Lambda}\right)^{1/\gamma}.
\end{equation}
Since $\varepsilon_T\le2\Lambda$, we have $h<1$. Set
\begin{equation}\label{eq:trunk-grid-sizes}
    n_t:=\left\lceil\frac{T}{h^2}\right\rceil,
    \qquad
    n_x:=\left\lceil\frac{2\pi}{h}\right\rceil,
    \qquad
    \tau:=\frac{T}{n_t},
    \qquad
    \sigma:=\frac{2\pi}{n_x}.
\end{equation}
The temporal grid $t_i=i\tau$, $0\le i\le n_t$, has spacing
$\tau\le h^2$, and the periodic spatial grid
$x_j=j\sigma$, $0\le j<n_x$, has spacing $\sigma\le h$.

For $0\le i\le n_t$, let $\vartheta_i$ be the continuous
piecewise-linear interpolation function on $[0,T]$ satisfying
\[
    \vartheta_i(t_k)=\delta_{ik},\qquad 0\le k\le n_t,
\]
and linear on each interval $[t_k,t_{k+1}]$. Similarly, for
$0\le j<n_x$, let $\psi_j$ be the continuous periodic
piecewise-linear interpolation function satisfying
\[
    \psi_j(x_k)=\delta_{jk},\qquad 0\le k<n_x.
\]
Here each function equals one at its own grid point and zero
at all the other grid points. The spatial functions are linear
on each grid interval, including $[x_{n_x-1},2\pi]$, with
$2\pi$ identified with $0$. These functions satisfy
\begin{equation}\label{eq:one-dimensional-pou}
    \vartheta_i,\psi_j\ge0,
    \qquad
    \sum_{i=0}^{n_t}\vartheta_i(t)=1,
    \qquad
    \sum_{j=0}^{n_x-1}\psi_j(x)=1.
\end{equation}
For $\alpha=(i,j_1,\ldots,j_d)$, define
\begin{equation}\label{eq:tensor-hat}
    z_\alpha:=(t_i,x_{j_1},\ldots,x_{j_d}),
    \qquad
    \varphi_\alpha(t,x)
       :=\vartheta_i(t)\prod_{q=1}^d\psi_{j_q}(x_q).
\end{equation}
Then $\varphi_\alpha\ge0$ and $\sum_\alpha\varphi_\alpha=1$.
Moreover, on the support of $\varphi_\alpha$, the temporal
displacement from $z_\alpha$ is at most $\tau$, and the spatial
distance is at most $\sqrt d\,\sigma$. Thus
\begin{equation}\label{eq:hat-support-distance}
    d_{\mathfrak s}(z,z_\alpha)
       \le \sqrt{\tau}+\sqrt d\,\sigma
       \le (1+\sqrt d)h.
\end{equation}

For $v\in\mathcal V$, define
\begin{equation}\label{eq:parabolic-interpolant}
    I_hv(z):=\sum_\alpha v(z_\alpha)\varphi_\alpha(z).
\end{equation}
The partition-of-unity property and the assumed H\"older bound give
\begin{align}
    |v(z)-I_hv(z)|
    &\le \sum_{\alpha:\,\varphi_\alpha(z)>0}
       \varphi_\alpha(z)|v(z)-v(z_\alpha)|\\
    &\le \Lambda((1+\sqrt d)h)^\gamma
     =\frac{\varepsilon_T}{2}.
    \label{eq:trunk-interpolation-error}
\end{align}
This estimate is uniform over $v\in\mathcal V$ and $z\in\DT$.

The number of grid nodes is
\begin{equation}\label{eq:trunk-node-count}
    N_1=(n_t+1)n_x^d.
\end{equation}
Since $h<1$ and the domain is fixed,
$n_t+1\asymp h^{-2}$ and $n_x\asymp h^{-1}$. Consequently,
\[
    N_1\asymp h^{-(d+2)}
       \asymp
       \left(\frac{\Lambda}{\varepsilon_T}\right)^{(d+2)/\gamma},
\]
which proves \eqref{eq:trunk-count}. The comparison constants
depend only on $T$, $d$, and $\gamma$.

\medskip
We next approximate the interpolation basis functions by ReLU networks.
The temporal interpolation functions have the explicit form
\[
    \vartheta_i(t)=\left(1-\frac{|t-t_i|}{\tau}\right)_+,
    \qquad t\in[0,T].
\]
On the coordinate interval $[0,2\pi]$, the periodic spatial
interpolation functions can be written as
\[
    \psi_j(x)=\sum_{k=-1}^{1}
       \left(1-\frac{|x-x_j-2\pi k|}{\sigma}\right)_+.
\]
By Lemma~\ref{lem:relu-primitives}(2), each of these
one-dimensional interpolation functions has an exact ReLU
representation with fixed depth and a fixed number of nonzero
parameters. Their values lie in $[0,1]$.
Each $\varphi_\alpha$ is a product of $d+1$ such factors.
The product construction in Lemma~\ref{lem:relu-primitives}
therefore gives, for every $\delta_T\in(0,1)$, a ReLU network
$\widetilde q_\alpha$ satisfying
\begin{equation}\label{eq:tensor-hat-network-error}
    \|\varphi_\alpha-\widetilde q_\alpha\|_{C(\DT)}
       \le\delta_T.
\end{equation}
Since $\psi_j(0)=\psi_j(2\pi)$ for every $j$, replacing any
spatial coordinate $x_q=0$ by $x_q=2\pi$, while keeping time
and the other coordinates fixed, leaves all inputs to the
product network unchanged. Thus $\widetilde q_\alpha$ takes
the same value at boundary points identified on $\T^d$ and
defines a continuous function on $\DT$. 
Relabel the nodes and networks by $\ell=1,\ldots,N_1$.
For every $v\in\mathcal V$, the interpolation estimate and
$|v(z_\ell)|\le B$ imply
\begin{align}
    \left\|v-\sum_{\ell=1}^{N_1}
          v(z_\ell)\widetilde q_\ell\right\|_{C(\DT)}
    &\le \|v-I_hv\|_{C(\DT)}
       +\sum_{\ell=1}^{N_1}|v(z_\ell)|
          \|\varphi_\ell-\widetilde q_\ell\|_{C(\DT)}
          \\
    &\le \frac{\varepsilon_T}{2}+BN_1\delta_T.
    \label{eq:trunk-total-decomposition}
\end{align}
Choose
\begin{equation}\label{eq:trunk-delta}
    \delta_T:=\frac{\varepsilon_T}{2BN_1}.
\end{equation}
Since $\varepsilon_T\le B$, this choice lies in $(0,1)$.
The preceding bound is then at most $\varepsilon_T$,
proving \eqref{eq:trunk-error}.

Furthermore, for every $z\in\DT$,
\begin{align*}
    \sum_{\ell=1}^{N_1}|\widetilde q_\ell(z)|
    &\le \sum_{\ell=1}^{N_1}\varphi_\ell(z)
       +\sum_{\ell=1}^{N_1}
          |\widetilde q_\ell(z)-\varphi_\ell(z)|\\
    &\le 1+N_1\delta_T
     =1+\frac{\varepsilon_T}{2B}
     \le\frac32.
\end{align*}
This proves \eqref{eq:trunk-l1}.

Finally, Lemma~\ref{lem:relu-primitives} gives each product
network depth and number of nonzero parameters bounded by
\[
    O\!\left((d+1)\log\frac{d+1}{\delta_T}\right).
\]
Combining this with the exact representations of the
interpolation functions and substituting \eqref{eq:trunk-delta}
gives $O((d+1)\log(BN_1/\varepsilon_T))$ for fixed $d$,
as claimed.
\end{proof}

\subsection{Functional approximation: the branch network}\label{sec:branch}

For $y\in\DT$, define the evaluation functional
\[
    \mathcal F_y(u_0,\zeta):=\cS(u_0,\zeta)(y).
\]
The forcing $\zeta$ is represented by its Fourier measurements
$c_J(\zeta)$, which are well defined for distribution-valued inputs.
Using the measurement map \eqref{eq:encoder}, we approximate the solution
value $\mathcal F_y(u_0,\zeta)$ as a function of the finite vector
$(\vec u_0,c_J(\zeta))$ by a partition of unity on a compact
coefficient set.  The
same partition weights are used for every $y$; only the solution values
$\Gamma_J(z^k)(y)$ depend on the output coordinate.

Fix $R\ge1$.  Assumption~\ref{ass:initial} gives
$|u_0(x_i)|\le M$ for every initial-data measurement.  On the event
$\{\|\eta\|_{X_0}\le R\}$, Lemma~\ref{lem:coeff} gives
\[
    \|c_J(\eta)\|_{\ell^\infty}
    \le C2^{-J\beta_0}R.
\]
We therefore set $R_J=C2^{-J\beta_0}R$ and define the compact box
\begin{equation}\label{eq:K}
    K_{J,R}:=[-M,M]^{N_0}\times[-R_J,R_J]^{N_J}.
\end{equation}
Then $(\vec u_0,c_J(\eta))\in K_{J,R}$ on this event.  The partition of unity
below is constructed on $K_{J,R}$.

We combine the reconstruction estimates with a partition of
unity on $K_{J,R}$ and ReLU approximation of its weights.

\begin{theorem}
\label{thm:branch}
Let $F:\R\to\R$ be globally Lipschitz, and let $\cS$ be
the corresponding solution map from Theorem~\ref{thm:stability}.
Suppose that Assumptions~\ref{ass:noise} and \ref{ass:initial}
hold, and fix $\beta_1$ and $\delta$ as in \eqref{eq:exponents}.
Let $R\ge1$ and $\varepsilon_B\in(0,1)$ be fixed. One can choose

\begin{equation}\label{eq:branch-choices}
    2^J\le C\left(\frac{R}{\varepsilon_B}\right)^{1/\delta},
    \qquad
    \delta_0=c_0\varepsilon_B,
    \qquad
    r=\frac{\varepsilon_B}{8L_J},
\end{equation}
where $c_0>0$ depends only on $d$, $T$, $L_F$, and $L_0$, and construct an $r/2$-net
$(z^k)_{k\le m}$ of $K_{J,R}$ and ReLU networks
$\widetilde\rho_k:\R^{N_0+N_J}\to\R$ such that
\begin{align}
&\sup_{u_0\in\cG_0}\sup_{\|\zeta\|_{X_0}\le R}
  \sup_{z_\bullet\in\DT}
  \left|
  \cS(u_0,\zeta)(z_\bullet)
  -\sum_{k=1}^m\Gamma_J(z^k)(z_\bullet)
       \widetilde\rho_k(\vec u_0,c_J(\zeta))
  \right|
  \le\varepsilon_B.\label{eq:branch-error}
\end{align}
The
networks can be chosen so that
\begin{equation}\label{eq:rho-l1}
    \sum_{k=1}^m
    |\widetilde\rho_k(z)|\le2,
    \qquad z\in\R^{N_0+N_J}.
\end{equation}
Moreover, the number $m$ of terms in the approximation can be chosen to
satisfy
\begin{equation}\label{eq:m-bound}
    m\le
    \left(1+\frac{2(M\vee R_J)\sqrt N}{r}\right)^N,
    \qquad N:=N_0+N_J.
\end{equation}
Choose a constant $C_B\ge1$, depending only on
$d,T,\beta_0,\beta_1,M,L_0,L_F$, and $|F(0)|$, so that
$B_J:=C_B(1+N_JR_J)$ bounds the reference solutions
$\Gamma_J(z^k)$ in $Y$, and set
\begin{equation}\label{eq:branch-complexity-log}
    H_B:=\log\!\left(\frac{eNm(B_J+1)}{\varepsilon_B}\right).
\end{equation}
Each $\widetilde\rho_k$ can be chosen with depth at most
$C\log(eN)H_B$ and at most $CNH_B$ nonzero weights and biases.
The vector-valued network
$\widetilde{\boldsymbol\rho}
 :=(\widetilde\rho_1,\ldots,\widetilde\rho_m)$ can be realized with
\begin{equation}\label{eq:partition-network-size}
    L_\rho\le C\log(eN)H_B,
    \qquad S_\rho\le CmNH_B,
\end{equation}
where $L_\rho$ denotes its depth and $S_\rho$ its total number of
nonzero weights and biases. These bounds include the projection onto
$K_{J,R}$; the constant $C$ is independent of $J$, $R$,
$\varepsilon_B$, $N$, and $m$.
\par
\end{theorem}

\begin{proof}
Fix $u_0\in\cG_0$ and $\zeta\in X_0$ with
$\|\zeta\|_{X_0}\le R$.  Introduce
\begin{equation}\label{eq:branch-reconstructed-data}
    \widehat u_0:=\Rec_0(\vec u_0),
    \qquad
    \zeta_J:=\Pi_J\zeta,
    \qquad
    z:=(\vec u_0,c_J(\zeta)).
\end{equation}
By \eqref{eq:sensors} and \eqref{eq:GammaJ},
\begin{equation}\label{eq:Gamma-at-encoded-data}
    \Gamma_J(z)=\cS(\widehat u_0,\zeta_J).
\end{equation}
Moreover, $z\in K_{J,R}$ by Assumption~\ref{ass:initial} and
\eqref{eq:coeff-bound}. For the Cartesian grid $(z^k)_{k\le m}$
and the tensor-product partition $(\rho_k)_{k\le m}$ constructed
in the estimate of $I_3$ below, set
\par
\begin{equation}\label{eq:AJr}
    A_{J,r}(z):=\sum_{k=1}^m\rho_k(z)\Gamma_J(z^k).
\end{equation}

We decompose the approximation error into four terms.
For any output point $z_\bullet\in\DT$, we have 
\begin{align}
&\left|
\cS(u_0,\zeta)(z_\bullet)
-\sum_{k=1}^m\Gamma_J(z^k)(z_\bullet)
       \widetilde\rho_k(z)
\right|\\
&\quad\le
\underbrace{\|\cS(u_0,\zeta)-\cS(u_0,\zeta_J)\|_Y}_{I_1}
+\underbrace{\|\cS(u_0,\zeta_J)-\Gamma_J(z)\|_Y}_{I_2}
\label{eq:four-errors}\\
&\qquad+
\underbrace{\|\Gamma_J(z)-A_{J,r}(z)\|_Y}_{I_3}
+\underbrace{\left|
\sum_{k=1}^m\Gamma_J(z^k)(z_\bullet)
       (\rho_k(z)-\widetilde\rho_k(z))
\right|}_{I_4}.
\end{align}
We estimate these terms separately.

\medskip
\noindent \textbf{Estimates of $I_1$.}
The two solutions in $I_1$ have the same initial datum, so the stability
estimate \eqref{eq:stability} and the cutoff estimate
\eqref{eq:projection} yield
\begin{equation}\label{eq:I1-bound}
    I_1
    \le {e^{L_FT}C_{\beta_1}}\|(I-\Pi_J)\zeta\|_{X_1}
    \le C_1 2^{-J\delta}\|\zeta\|_{X_0}
    \le C_1 2^{-J\delta}R,
\end{equation}
where $C_1\ge1$ is independent of $J$, $R$, and $\zeta$.

We choose $J$ to be
the smallest nonnegative integer satisfying
\begin{equation}\label{eq:J-minimal-choice}
    2^{J\delta}\ge\frac{4C_1R}{\varepsilon_B}.
\end{equation}
Then $I_1\le\varepsilon_B/4$. Since $4C_1R/\varepsilon_B>1$, we have $J\ge1$.
By the choice of $J$, the preceding integer $J-1$ does
not satisfy \eqref{eq:J-minimal-choice}. Thus
\[
    2^{(J-1)\delta}<\frac{4C_1R}{\varepsilon_B},
\]
and consequently
\[
    2^J<2\left(\frac{4C_1R}{\varepsilon_B}\right)^{1/\delta},
\]
which is the first estimate in \eqref{eq:branch-choices}.  Taking the
smallest $J$ is important because both $N_J$ and the Lipschitz
constant $L_J$ increase with the cutoff level.

\medskip
\noindent\textbf{Estimates of $I_2$.}
By \eqref{eq:Gamma-at-encoded-data}, the forcing in the two solutions
defining $I_2$ is the same.  Hence \eqref{eq:stability} and
\eqref{eq:appendix-interpolation} give
\begin{equation}\label{eq:I2-bound}
    I_2
    =\|\cS(u_0,\zeta_J)-\cS(\widehat u_0,\zeta_J)\|_Y
    \le {e^{L_FT}}\|u_0-\widehat u_0\|_\infty
    \le C_2L_0\delta_0.
\end{equation}
{
Here $C_2:=e^{L_FT}C_d$, where $C_d$ is the interpolation constant
in \eqref{eq:appendix-interpolation}.
\par}
Choose

\begin{equation}\label{eq:delta0-choice}
    \delta_0=c_0\varepsilon_B,
    \qquad c_0:=(4C_2L_0)^{-1}.
\end{equation}
Then $I_2\le\varepsilon_B/4$, and the number of
initial-data measurements satisfies
\[
    N_0\asymp\delta_0^{-d}\asymp\varepsilon_B^{-d}.
\]

\medskip
\noindent\textbf{Estimates of $I_3$.}
For $z_1=(\vec v,c)$ and $z_2=(\vec v',c')$, one has
\begin{align*}
|\vec v-\vec v'|_{\ell^\infty}+|c-c'|_{\ell^2}
&\le |\vec v-\vec v'|_{\ell^2}+|c-c'|_{\ell^2}\\
&\le2|z_1-z_2|_{\ell^2}.
\end{align*}
Thus Lemma~\ref{lem:Gamma-lip} shows that $\Gamma_J$ is
$2L_J$-Lipschitz on the Euclidean box $K_{J,R}$.
We construct an $r/2$-net of this box as follows. Set $N=N_0+N_J$. For $a\in\{M,R_J\}$, define
\[
    n_a:=\left\lceil\frac{2a\sqrt N}{r}\right\rceil,
    \qquad
    G_a:=\left\{-a+\frac{(2j-1)a}{n_a}:
                   j=1,\ldots,n_a\right\}.
\]
Let
\[
    \{z^1,\ldots,z^m\}:=G_M^{N_0}\times G_{R_J}^{N_J}
    \subset K_{J,R}.
\]
For every $z\in K_{J,R}$, one can choose a center $z^k$
such that
\[
    |z-z^k|_{\ell^\infty}\le\frac{r}{2\sqrt N}.
\]
Consequently,
\[
    |z-z^k|_{\ell^2}
    \le\sqrt N\,|z-z^k|_{\ell^\infty}
    \le\frac r2,
\]
so these centers form an $r/2$-net of $K_{J,R}$. Moreover,
\[
    m=n_M^{N_0}n_{R_J}^{N_J}
    \le
    \left(1+\frac{2(M\vee R_J)\sqrt N}{r}\right)^N,
\]
which proves \eqref{eq:m-bound}.
We now define interpolation weights on this grid. Write
$G_a=\{\xi_{a,j}:1\le j\le n_a\}$ in increasing order and set
$s_a:=2a/n_a\le r/\sqrt N$. If $n_a=1$, set $b_{a,1}\equiv1$.
If $n_a\ge2$, define
\begin{align*}
    b_{a,1}(s)
      &:=P\!\left(\frac{\xi_{a,2}-s}{s_a}\right),\\
    b_{a,n_a}(s)
      &:=P\!\left(\frac{s-\xi_{a,n_a-1}}{s_a}\right),\\
    b_{a,j}(s)
      &:=\left(1-\frac{|s-\xi_{a,j}|}{s_a}\right)_+,
      \qquad 1<j<n_a,
\end{align*}
where
\[
    P(v):=
    \begin{cases}
        0,&v<0,\\
        v,&0\le v\le1,\\
        1,&v>1.
    \end{cases}
\]
These functions interpolate the grid values and are extended constantly
between the first or last grid point and the corresponding boundary.
For $s\in[-a,a]$, they satisfy
\[
    b_{a,j}(s)\ge0,\qquad
    \sum_{j=1}^{n_a}b_{a,j}(s)=1,\qquad
    b_{a,j}(s)>0\ \Longrightarrow\ |s-\xi_{a,j}|\le s_a.
\]
The last bound also holds when $n_a=1$, since then $\xi_{a,1}=0$
and $|s|\le a\le s_a$.

Set $a_q=M$ for $1\le q\le N_0$ and $a_q=R_J$ for
$N_0<q\le N$. Index the grid points by
$\alpha=(j_1,\ldots,j_N)$ and define
\begin{equation}\label{eq:branch-tensor-weights}
    z^\alpha:=(\xi_{a_q,j_q})_{q=1}^N,
    \qquad
    \rho_\alpha(w):=\prod_{q=1}^N b_{a_q,j_q}(w_q).
\end{equation}
For each center $z^k=z^\alpha$, set $\rho_k:=\rho_\alpha$.
On $K_{J,R}$, these weights are nonnegative and sum to one.
Moreover, for $w\in K_{J,R}$,
\[
    \rho_k(w)>0\quad\Longrightarrow\quad
    |w-z^k|_{\ell^2}
       \le\left(\sum_{q=1}^N s_{a_q}^2\right)^{1/2}\le r.
\]
The $2L_J$-Lipschitz continuity of $\Gamma_J$ therefore gives
\begin{equation}\label{eq:I3-bound}
    I_3
    \le\sum_{k=1}^m\rho_k(z)
       \|\Gamma_J(z)-\Gamma_J(z^k)\|_Y
    \le2L_Jr.
\end{equation}

Choosing
\begin{equation}\label{eq:r-choice}
    r:=\frac{\varepsilon_B}{8L_J}
\end{equation}
therefore gives $I_3\le\varepsilon_B/4$.

\medskip
\noindent\textbf{Estimates of $I_4$.}
Write $z^k=(\vec v^{\,k},c^k)$. Since $z^k\in K_{J,R}$,
each entry $v_i^{\,k}$ belongs to $[-M,M]$.
At every spatial point, $\Rec_0(\vec v^{\,k})$ is a
weighted average of these grid values, with nonnegative
weights summing to one. Hence
\[
    \|\Rec_0(\vec v^{\,k})\|_\infty\le M.
\]
Similarly, $|c^k|_{\ell^2}\le\sqrt{N_J}R_J$, and
\eqref{eq:rec-bound} gives
\[
    \|\Rec_J(c^k)\|_{X_1}
    \le C N_J^{1/2}|c^k|_{\ell^2}
    \le C N_JR_J.
\]
The a priori estimate \eqref{eq:apriori} now implies, for a
constant $C_B\ge1$ depending only on the parameters specified above,
\begin{equation}\label{eq:BJ}
    \max_{k\le m}\|\Gamma_J(z^k)\|_Y
    \le B_J:=C_B\big(1+N_JR_J\big).
\end{equation}
Each one-dimensional function $b_{a,j}$ has an exact ReLU
representation of fixed depth with a fixed number of nonzero parameters:
the endpoint functions use Lemma~\ref{lem:relu-primitives}(1),
the interior functions use part~(2), and the case $n_a=1$ is constant.
All their values lie in $[0,1]$. Each weight in
\eqref{eq:branch-tensor-weights} is a product of $N$ such factors.
The product construction in Lemma~\ref{lem:relu-primitives} therefore
gives, for every $\delta_B\in(0,1)$, ReLU networks $\widehat\rho_k$
with values in $[0,1]$ such that
\begin{equation}\label{eq:rho-uniform-approximation}
    \max_{k\le m}
    \|\rho_k-\widehat\rho_k\|_{C(K_{J,R})}
    \le\delta_B.
\end{equation}
For $N=1$, the factor itself gives an exact representation.
In all cases, each network $\widehat\rho_k$ can be realized with
\begin{equation}\label{eq:weight-network-size}
\begin{aligned}
    \operatorname{depth}(\widehat\rho_k)
       &\le C\log(eN)\log\frac{eN}{\delta_B},\\
    \operatorname{size}(\widehat\rho_k)
       &\le CN\log\frac{eN}{\delta_B},
\end{aligned}
\end{equation}
where size denotes the total number of nonzero weights and biases.
The $m$ networks can be realized at the same depth, using
the same product construction and fixed-depth representations of their
one-dimensional factors. The projections onto $[0,1]$ in the product construction ensure
that the outputs belong to $[0,1]$.
\par
Let $P_{K_{J,R}}$ map each coordinate to the nearest point
in its defining interval in $K_{J,R}$. This map leaves
every point of $K_{J,R}$ unchanged and maps every input
into $K_{J,R}$. Define
\begin{equation}\label{eq:projected-rho}
    \widetilde\rho_k:=\widehat\rho_k\circ P_{K_{J,R}}.
\end{equation}
Lemma~\ref{lem:relu-primitives}(1) shows that this is again a ReLU network.
Since $P_{K_{J,R}}$ is the identity on $K_{J,R}$,
\eqref{eq:rho-uniform-approximation} gives, for the last term in
\eqref{eq:four-errors},
\begin{align}
    I_4
    &\le\sum_{k=1}^m
       \|\Gamma_J(z^k)\|_Y
       |\rho_k(z)-\widetilde\rho_k(z)|\\
    &\le mB_J\delta_B.
    \label{eq:I4-bound}
\end{align}
Set
\begin{equation}\label{eq:deltaB-choice}
    \delta_B:=\frac{\varepsilon_B}{4m(B_J+1)}.
\end{equation}
Then $I_4\le\varepsilon_B/4$.  The same choice gives, for every
$w\in\R^{N_0+N_J}$,
\begin{align*}
\sum_{k=1}^m|\widetilde\rho_k(w)|
&\le\sum_{k=1}^m\rho_k(P_{K_{J,R}}w)+m\delta_B\\
&\le1+\frac{\varepsilon_B}{4(B_J+1)}
\le2,
\end{align*}
which proves \eqref{eq:rho-l1}.
For the choice \eqref{eq:deltaB-choice},
\[
    \log\frac{eN}{\delta_B}
    =\log\frac{4eNm(B_J+1)}{\varepsilon_B}\le C H_B.
\]
The coordinate projection requires $O(N)$ nonzero parameters and fixed
depth. Thus \eqref{eq:weight-network-size} gives the stated bounds for
each $\widetilde\rho_k$. Running these networks in parallel gives
\eqref{eq:partition-network-size}. Since their outputs are nonnegative,
their final affine outputs may be treated as ReLU hidden activations.
Appending an identity output map adds at most $m$ parameters and one
layer, which are absorbed by the same bounds.
\par

Combining the estimates from \eqref{eq:J-minimal-choice}, \eqref{eq:delta0-choice},
\eqref{eq:r-choice}, and \eqref{eq:deltaB-choice}, give
$I_i\le\varepsilon_B/4$ for $1\le i\le4$.  Summing
\eqref{eq:four-errors} proves \eqref{eq:branch-error}, uniformly in
$u_0$, $\zeta$, and $z_\bullet$.
\end{proof}

\subsection{Operator approximation}\label{sec:main}

We now combine the function approximation of
Theorem~\ref{thm:trunk} with the functional approximation of
Theorem~\ref{thm:branch}.  This is the SPDE counterpart of the final operator
approximation step in \cite{bayraktar2025deep}.

\begin{theorem}
\label{thm:main}
Let $d\ge2$, let Assumptions~\ref{ass:noise} and \ref{ass:initial} hold, and
let $F$ be globally Lipschitz.  Choose $\beta_1$ as in
\eqref{eq:exponents} and $\gamma$ as in \eqref{eq:gamma}.  For every
$\varepsilon\in(0,1)$ there are a cutoff level $J$, $N_0$ initial-data
measurements, and ReLU branch and trunk networks such that
\begin{equation}\label{eq:main}
    \sup_{u_0\in\cG_0}
    \E\left[
      \sup_{(t,x)\in\DT}
      \left|
      \cS(u_0,\eta)(t,x)
      -\sum_{\ell=1}^{N_1}
       \widetilde a_\ell(\vec u_0,c_J(\eta))
       \widetilde q_\ell(t,x)
      \right|
    \right]
    \le\varepsilon.
\end{equation}
The parameters may be chosen so that, with
$R=C_0\sqrt{\log(e/\varepsilon)}$,
\begin{align}
    N_J&\le C\left(\frac{R}{\varepsilon}\right)^{(d+2)/\delta},
        \label{eq:NJ-rate}\\
    N_0&\le C\varepsilon^{-d},\label{eq:N0-rate}\\
    N_1&\le C\left(\frac{1+R}{\varepsilon}\right)^{(d+2)/\gamma}.
        \label{eq:N1-rate}
\end{align}
For each trunk network, its depth $L_{T,\ell}$ and number
$S_{T,\ell}$ of nonzero weights and biases satisfy
\begin{equation}\label{eq:main-trunk-size}
    L_{T,\ell}\le C\log(e/\varepsilon),
    \qquad S_{T,\ell}\le C\log(e/\varepsilon),
    \qquad 1\le\ell\le N_1.
\end{equation}
The branch networks can be realized as the coordinates of a single
network $\widetilde{\boldsymbol a}:\R^N\to\R^{N_1}$, where
$N=N_0+N_J$. This network uses $m$ shared partition weights, with
\begin{equation}\label{eq:main-feature-count}
    m\le\left(1+\frac{2(M\vee R_J)\sqrt N}{r}\right)^N,
    \qquad r=\frac{\varepsilon}{48L_J}.
\end{equation}
Set $B_J:=C_B(1+N_JR_J)$, with $C_B$ as in
Theorem~\ref{thm:branch}, and define
\begin{equation}\label{eq:main-complexity-log}
    H_\varepsilon
    :=\log\!\left(\frac{eNm(B_J+1)}{\varepsilon}\right).
\end{equation}
The shared branch network has depth $L_B$ and total number $S_B$
of nonzero weights and biases bounded by
\begin{equation}\label{eq:main-branch-size}
    L_B\le C\log(eN)H_\varepsilon,
    \qquad S_B\le C\bigl(mNH_\varepsilon+mN_1\bigr).
\end{equation}
Here the shared partition weights are counted once, and $mN_1$
counts the coefficients in the final linear map. Consequently, the
combined number of nonzero parameters in the branch and trunk networks is
\begin{equation}\label{eq:main-total-size}
    S_B+\sum_{\ell=1}^{N_1}S_{T,\ell}
    \le C\bigl(mNH_\varepsilon+mN_1
               +N_1\log(e/\varepsilon)\bigr).
\end{equation}
The constants in these size bounds are independent of $\varepsilon$.
\par
\end{theorem}

\begin{proof}
{\textbf{Construction and uniform approximation.}}
For a radius $R\ge1$, to be chosen below, define
\begin{equation}\label{eq:good-event}
    \Omega_R:=\{\|\eta\|_{X_0}\le R\}.
\end{equation}
We first construct an approximation whose error on $\Omega_R$ is
uniformly bounded by $\varepsilon/2$.

Let $\varepsilon_T$ denote the error in reconstructing an output function
from the trunk nodes, and let $\varepsilon_B$ denote the error in
approximating its value at each trunk node.  The latter is multiplied by the
$\ell^1$ norm of the trunk basis.  In view of \eqref{eq:trunk-l1}, it is
enough to require
\begin{equation}\label{eq:operator-error-budget}
    \varepsilon_T+\frac32\varepsilon_B
       \le\frac\varepsilon2.
\end{equation}
We use the convenient choice
\begin{equation}\label{eq:operator-eps-choices}
    \varepsilon_T:=\frac\varepsilon4,
    \qquad
    \varepsilon_B:=\frac\varepsilon6,
\end{equation}
for which equality holds in \eqref{eq:operator-error-budget}.

Consider the class of output functions generated on the noise ball,
\begin{equation}\label{eq:VR}
    \mathcal V_R
    :=\{\cS(u_0,\zeta):u_0\in\cG_0,
          \ \zeta\in X_0,\ \|\zeta\|_{X_0}\le R\}.
\end{equation}
Lemma~\ref{lem:modulus} gives, for every $v\in\mathcal V_R$,
\[
    \|v\|_\infty\le B(R),
    \qquad
    |v(y)-v(y')|
       \le\Lambda(R)d_{\mathfrak s}(y,y')^\gamma.
\]
Apply Theorem~\ref{thm:trunk} to $\mathcal V_R$ with the accuracy
$\varepsilon_T$ in \eqref{eq:operator-eps-choices}; the hypothesis
$\varepsilon_T\le2\Lambda(R)$ holds because $\varepsilon_T<1\le\Lambda(R)$,
and the companion condition $\varepsilon_T\le B(R)$ follows from $B(R)\ge1$.
We obtain nodes $y_1,\ldots,y_{N_1}$ and trunk networks
$\widetilde q_1,\ldots,\widetilde q_{N_1}$ satisfying
\begin{align}
    \sup_{v\in\mathcal V_R}
    \left\|v-\sum_{\ell=1}^{N_1}
       v(y_\ell)\widetilde q_\ell\right\|_Y
       &\le\varepsilon_T,
       \label{eq:operator-trunk-approximation}\\
    \sup_{y\in\DT}\sum_{\ell=1}^{N_1}
       |\widetilde q_\ell(y)|&\le\frac32.
       \label{eq:operator-trunk-l1}
\end{align}
The nodes and the trunk networks depend on $R$ and $\varepsilon$, but they do
not depend on a particular $u_0$ or realization of the forcing.

Next, apply Theorem~\ref{thm:branch} with the same radius $R$ and with accuracy
$\varepsilon_B$ from \eqref{eq:operator-eps-choices}.  This determines
$J$, $\delta_0$, the compact coefficient box $K_{J,R}$, centers
$z^1,\ldots,z^m$, and common partition networks
$\widetilde\rho_1,\ldots,\widetilde\rho_m$.  The construction is uniform in
the output point.  In particular, for every $y\in\DT$,
\begin{equation}\label{eq:operator-functional-uniform}
\sup_{u_0\in\cG_0}\sup_{\|\zeta\|_{X_0}\le R}
\left|
\cS(u_0,\zeta)(y)
-\sum_{k=1}^m\Gamma_J(z^k)(y)
 \widetilde\rho_k(\vec u_0,c_J(\zeta))
\right|
\le\varepsilon_B.
\end{equation}

For each trunk node $y_\ell$, define the branch network
\begin{equation}\label{eq:branch-readout}
    \widetilde a_\ell(w)
    :=\sum_{k=1}^m
       \Gamma_J(z^k)(y_\ell)\widetilde\rho_k(w),
    \qquad w\in\R^{N_0+N_J}.
\end{equation}
These branch networks can be combined into a single vector-valued network:
\[
    \widetilde{\boldsymbol a}(w)
       =A\widetilde{\boldsymbol\rho}(w),
    \qquad A_{\ell k}:=\Gamma_J(z^k)(y_\ell),
    \quad A\in\R^{N_1\times m}.
\]
Theorem~\ref{thm:branch} realizes the shared weights at a common depth
with nonnegative outputs as the final hidden activations. Appending
the linear map $A$ therefore gives a ReLU network with at most
$mN_1$ additional nonzero parameters and zero output bias.
Evaluating
\eqref{eq:operator-functional-uniform} at $y=y_\ell$ yields
\begin{equation}\label{eq:branch-node-error}
    \max_{\ell\le N_1}
    |\cS(u_0,\zeta)(y_\ell)
      -\widetilde a_\ell(\vec u_0,c_J(\zeta))|
    \le\varepsilon_B
\end{equation}
whenever $u_0\in\cG_0$ and $\|\zeta\|_{X_0}\le R$.

We now combine these approximations.
Fix $u_0\in\cG_0$ and a forcing realization in $\Omega_R$, and write
$u:=\cS(u_0,\eta)$.  For every $y\in\DT$, add and subtract the trunk
interpolant with the exact nodal values:
\begin{align}
&\left|u(y)-\sum_{\ell=1}^{N_1}
   \widetilde a_\ell(\vec u_0,c_J(\eta))
   \widetilde q_\ell(y)\right|\\
&\quad\le
\left|u(y)-\sum_{\ell=1}^{N_1}
       u(y_\ell)\widetilde q_\ell(y)\right|
\label{eq:operator-two-errors}\\
&\qquad+
\sum_{\ell=1}^{N_1}
  |u(y_\ell)-\widetilde a_\ell(\vec u_0,c_J(\eta))|
  |\widetilde q_\ell(y)|.
\end{align}
The first term is bounded by \eqref{eq:operator-trunk-approximation}; the
second is bounded using \eqref{eq:branch-node-error} and
\eqref{eq:operator-trunk-l1}.  Taking the supremum over $y$ gives
\begin{align}
\left\|u-\sum_{\ell=1}^{N_1}
   \widetilde a_\ell(\vec u_0,c_J(\eta))
   \widetilde q_\ell\right\|_Y
&\le\varepsilon_T+\frac{3}{2}\varepsilon_B
\le\frac\varepsilon2.
\label{eq:good-event-operator-error}
\end{align}
This is the deterministic branch--trunk approximation on $\Omega_R$.

\medskip
{\bf Expected approximation error.}
The {coordinatewise projection} in \eqref{eq:projected-rho} ensures that the branch
network remains bounded even when the encoded forcing lies outside
$K_{J,R}$.  Indeed, \eqref{eq:BJ}, \eqref{eq:rho-l1}, and
\eqref{eq:branch-readout} give, for every $w\in\R^{N_0+N_J}$,
\begin{equation}\label{eq:branch-global-bound}
    |\widetilde a_\ell(w)|
    \le B_J\sum_{k=1}^m|\widetilde\rho_k(w)|
    \le2B_J.
\end{equation}
Combining this with \eqref{eq:operator-trunk-l1},
\begin{equation}\label{eq:network-global-bound}
    \sup_{y\in\DT}
    \left|\sum_{\ell=1}^{N_1}
       \widetilde a_\ell(w)\widetilde q_\ell(y)\right|
    \le3B_J.
\end{equation}

We next express $B_J$ in terms of $R$ and $\varepsilon$.  From
\eqref{eq:BJ}, the definition of $R_J$, and \eqref{eq:rank},
\begin{align*}
    B_J
    &\le C\big(1+N_J2^{-J\beta_0}R\big)\\
    &\le C\big(1+R2^{J(d+2-\beta_0)}\big).
\end{align*}
Since $\beta_0>-2$, one has
$d+2-\beta_0<d+4$.  The choice of $J$ in
Theorem~\ref{thm:branch}, with $\varepsilon_B=\varepsilon/6$, therefore
implies
\begin{equation}\label{eq:BJ-poly}
    B_J\le C\left[1+R
      \left(\frac{R}{\varepsilon}\right)^{(d+4)/\delta}
    \right].
\end{equation}

Let
\begin{equation*}
    \mathcal E_\theta(u_0,\eta)
    :=\left\|\cS(u_0,\eta)
      -\sum_{\ell=1}^{N_1}
       \widetilde a_\ell(\vec u_0,c_J(\eta))
       \widetilde q_\ell\right\|_Y.
\end{equation*}
For each fixed $u_0\in\cG_0$, the map
$\eta\mapsto\mathcal E_\theta(u_0,\eta)$ is Borel measurable.  Indeed, the
measurement map is continuous on $X_0$, the solution map $\cS$ is Lipschitz from
$C(\T^d)\times X_1$ to the separable space $Y$, the embedding
$X_0\hookrightarrow X_1$ is continuous, and the norm on $Y$ is continuous.
Thus the expectations below are well defined.
The a priori estimate \eqref{eq:apriori} and
\eqref{eq:network-global-bound} imply
\begin{equation}\label{eq:error-global-growth}
    \mathcal E_\theta(u_0,\eta)
    \le C\big(1+B_J+\|\eta\|_{X_0}\big),
\end{equation}
uniformly in $u_0\in\cG_0$.  Set
$p:=(d+4)/\delta+1$.  By \eqref{eq:BJ-poly}, after increasing $C$,
\[
    1+B_J\le C\varepsilon^{-p}(1+R^p).
\]

Lemmas~\ref{lem:fernique} and \ref{lem:fernique}, applied to
\eqref{eq:error-global-growth}, now give constants $c_1,C>0$ such that
\begin{equation}\label{eq:bad-event-bound}
\sup_{u_0\in\cG_0}
\E[\mathcal E_\theta(u_0,\eta)\one_{\Omega_R^c}]
\le C\varepsilon^{-p}(1+R^p)e^{-c_1R^2}.
\end{equation}
The constants \(C,c_1\) are independent of \(R\ge1\). We now choose
\begin{equation}\label{eq:R-choice}
    R:=C_0\sqrt{\log(e/\varepsilon)}.
\end{equation}
The right-hand side of \eqref{eq:bad-event-bound} is bounded by
\[
    C\varepsilon^{-p}
    \big(1+C_0^p\log(e/\varepsilon)^{p/2}\big)
    (e/\varepsilon)^{-c_1C_0^2}.
\]
Taking $C_0$ sufficiently large, depending only on the fixed data, makes
$c_1C_0^2>p+2$ and absorbs the remaining constant and logarithmic factor.
The expression is then at most $\varepsilon/2$ for
$\varepsilon\in(0,1)$.  Together with
\eqref{eq:good-event-operator-error}, this yields
\begin{align*}
\sup_{u_0\in\cG_0}\E[\mathcal E_\theta(u_0,\eta)]
&\le\frac\varepsilon2\PP(\Omega_R)
  +\sup_{u_0\in\cG_0}
   \E[\mathcal E_\theta(u_0,\eta)\one_{\Omega_R^c}]\\
&\le\varepsilon,
\end{align*}
which proves \eqref{eq:main}.

\medskip
\noindent{\textbf{Measurement counts and network sizes.}}
With $\varepsilon_B=\varepsilon/6$, the minimal cutoff choice
\eqref{eq:J-minimal-choice} gives
\[
    2^J\le C\left(\frac{R}{\varepsilon}\right)^{1/\delta}.
\]
Using $N_J\asymp2^{J(d+2)}$ proves \eqref{eq:NJ-rate}.  Likewise,
\eqref{eq:delta0-choice} gives $\delta_0\asymp\varepsilon$, and the periodic
grid count $N_0\asymp\delta_0^{-d}$ proves \eqref{eq:N0-rate}.  Finally,
Lemma~\ref{lem:modulus} gives $\Lambda(R)\le C(1+R)$, so
\eqref{eq:trunk-count} with $\varepsilon_T=\varepsilon/4$ proves
\eqref{eq:N1-rate}.

{
For each trunk network, Theorem~\ref{thm:trunk} bounds its depth and
number of nonzero parameters by
\[
    C\log\!\left(\frac{4B(R)N_1}{\varepsilon}\right).
\]
Since $B(R)\le C(1+R)$, \eqref{eq:N1-rate} implies that this is at most
$C\log(e(1+R)/\varepsilon)$, with $d$ and $\gamma$ fixed.
The choice $R=C_0\sqrt{\log(e/\varepsilon)}$ then gives
\eqref{eq:main-trunk-size}.

For the shared branch network, the choice
$\varepsilon_B=\varepsilon/6$ gives
$r=\varepsilon/(48L_J)$ and hence \eqref{eq:main-feature-count}.
It also gives
\[
    H_B=\log\!\left(\frac{6eNm(B_J+1)}{\varepsilon}\right)
       \le C H_\varepsilon.
\]
Thus \eqref{eq:partition-network-size} bounds the depth and number
of parameters in the shared feature network by
$C\log(eN)H_\varepsilon$ and $CmNH_\varepsilon$, respectively.
The final linear map in \eqref{eq:branch-readout} adds at most $mN_1$
nonzero parameters and one layer. This proves
\eqref{eq:main-branch-size}. Summing the branch and trunk parameter
counts proves \eqref{eq:main-total-size}.
\par}
\end{proof}

\begin{corollary}
\label{cor:Lp-approximation}
Under the assumptions of Theorem~\ref{thm:main}, let $q\in[1,\infty)$.  For
every $\varepsilon\in(0,1)$ there are finite measurement maps and ReLU branch and trunk
networks such that
\begin{equation}\label{eq:Lp-approximation}
\sup_{u_0\in\cG_0}
\left(
\E\left[
  \left\|
  \cS(u_0,\eta)
  -\sum_{\ell=1}^{N_1}
   \widetilde a_\ell(\vec u_0,c_J(\eta))
   \widetilde q_\ell
  \right\|_Y^q
\right]
\right)^{1/q}
\le\varepsilon.
\end{equation}
The measurement and trunk counts satisfy
\eqref{eq:NJ-rate}--\eqref{eq:N1-rate}, and the network depths
and parameter counts satisfy \eqref{eq:main-trunk-size},
\eqref{eq:main-branch-size}, and \eqref{eq:main-total-size}, with
\[
    R=C_q'\sqrt{\log(e/\varepsilon)}.
\]
Here $C_q'\ge1$, the quantities $N$, $m$, $B_J$, and
$H_\varepsilon$ are defined as in Theorem~\ref{thm:main} for
this choice of $R$, and the constants may depend on $q$.
\par
\end{corollary}

\begin{proof}
Use the construction in Theorem~\ref{thm:main}, with the deterministic error
on $\Omega_R$ bounded by $\varepsilon/2$, and denote the resulting random error by
$\mathcal E_\theta$.  Then
\[
    \mathcal E_\theta\one_{\Omega_R}\le\frac\varepsilon2.
\]
On $\Omega_R^c$, the {coordinatewise projection onto $K_{J,R}$},
the bound $B_J$ for the reference solutions,
and the a priori estimate give
\[
    \mathcal E_\theta
    \le C\big(B_J+1+\|\eta\|_{X_0}\big).
\]
For $p_0:=(d+4)/\delta+1$, equations \eqref{eq:BJ-poly} and
\eqref{eq:branch-choices} imply
$B_J+1\le C\varepsilon^{-p_0}(1+R^{p_0})$.
Since $B_J+1\ge1$, the deterministic factorization
$(B_J+1+\|\eta\|_{X_0})^q\le(B_J+1)^q(1+\|\eta\|_{X_0})^q$, together with
Lemma~\ref{lem:fernique} at exponent $q$ and
$(1+R^{p_0})^q\le2^q(1+R^{qp_0})$, therefore yields
\[
    \E\left[\mathcal E_\theta^q\one_{\Omega_R^c}\right]
    \le C_q\varepsilon^{-qp_0}(1+R^{qp_0})e^{-c_qR^2}.
\]
Choose $R=C_q'\sqrt{\log(e/\varepsilon)}$ with $C_q'$ large enough that the
right-hand side is at most $(\varepsilon/2)^q$.  Adding the good- and bad-event
contributions and taking the $q$th root gives the bound
$2^{1/q}\varepsilon/2\le\varepsilon$, since $q\ge1$; this proves
\eqref{eq:Lp-approximation}.
With this choice of $R$, the measurement and network size estimates
follow by the same calculations as in Theorem~\ref{thm:main}, with
constants that may depend on $q$.
\par
\end{proof}

\section{Local approximation for stochastic Navier--Stokes equations}
\label{sec:nse}

We consider the stochastic incompressible Navier--Stokes equations on
$\T^d$, $d\ge2$. In contrast to the globally Lipschitz nonlinearity in
Section~\ref{sec:setting}, the transport term is quadratic and contains
a spatial derivative. We establish local existence and Lipschitz
stability on a common interval for a bounded set of initial velocities
and forcing convolutions. The Fourier measurement maps and partition of unity
from Section~\ref{sec:reduction} then give an approximation uniform on
this set. Gaussian tail estimates determine the probability that the
random forcing satisfies the required bound.

\subsection{The stochastic equation and solution spaces}

Fix $\nu>0$. Consider
\begin{equation}\label{eq:nse}
\begin{cases}
    \partial_tu+(u\cdot\nabla)u+\nabla p=\nu\Delta u+\eta,
       &(t,x)\in(0,T]\times\T^d,\\
    \nabla\cdot u=0,\qquad u(0)=u_0.
\end{cases}
\end{equation}
Here $\eta$ is a centered Gaussian Borel random variable with values in
$(\cC_g^{\beta_0}(\Lambda_T))^d$, as in
Assumption~\ref{ass:noise}, interpreted for vector-valued forcing.
Its realizations may be distributions rather than functions.
Equation~\eqref{eq:nse} is interpreted pathwise, after restriction
of the forcing to $D_T$.
Both $u_0$ and the forcing are assumed to be divergence-free and to have
zero spatial mean; for $\eta$, these conditions hold almost surely.
For a distributional forcing, these conditions are
understood in the sense of distributions. Let $P$ be the Leray projector
defined in Section~\ref{sec:pre-operators}. For deterministic forcing $\phi$, the mild formulation is
\begin{equation}\label{eq:nse-mild}
\begin{split}
    u(t)&=e^{\nu t\Delta}u_0+K_\nu\phi(t)-N(u,u)(t),\\
    N(u,v)(t)&:=\int_0^t e^{\nu(t-\sigma)\Delta}
                  P\nabla\cdot(u\otimes v)(\sigma)\,d\sigma.
\end{split}
\end{equation}
We will apply this formulation both to realizations of $\eta$ and
to their Fourier approximations. Fix $s\in(0,1)$ and define
\[
    \cC_x^s:=B^s_{\infty,\infty}(\T^d;\R^d),
    \qquad
    \cC_{g,x}^s:=\overline{C^\infty(\T^d;\R^d)}^{\,\cC_x^s}.
\]
The local-in-time solution space is
\[
    E_\tau:=\left\{u\in C([0,\tau];\cC_x^s):
       \nabla\cdot u(t)=0,\ \int_{\T^d}u(t,x)\,dx=0\right\},
    \qquad
    \|u\|_{E_\tau}:=\sup_{t\le\tau}\|u(t)\|_{\cC_x^s}.
\]
We distinguish this space from the approximation space
\begin{equation}\label{eq:nse-output-space}
    Y_\tau:=C([0,\tau]\times\T^d;\R^d),
    \qquad \|u\|_{Y_\tau}:=\sup_{(t,x)}|u(t,x)|.
\end{equation}
Here $C_{\rm emb}$ denotes a constant such that
$\|u\|_{Y_\tau}\le C_{\rm emb}\|u\|_{E_\tau}$, uniformly for $\tau\le T$.
All norms of vector-valued functions use the maximum over components. Taking the closure of smooth functions ensures continuity of the free
solution at the initial time in the $\cC_x^s$ norm. The heat semigroup is strongly
continuous on $\cC_{g,x}^s$, but not on the whole of $\cC_x^s$.
The initial-data class used in the approximation theorem lies in
$\cC_x^{\bar s}$ for $\bar s>s$, and is therefore contained in
$\cC_{g,x}^s$.

\subsection{Local existence and stability on a common interval}

We present the estimates needed for the fixed-point argument. The spatial
Littlewood--Paley norms are those of Section~\ref{sec:pre-operators}.

\begin{lemma}[Product, projection, and semigroup estimates]
\label{lem:nse-algebra}
Let $0<s<1$. There is a constant $C_s$, depending only on $d$ and $s$,
such that
\begin{align*}
    \|fg\|_{\cC_x^s}&\le C_s\|f\|_{\cC_x^s}\|g\|_{\cC_x^s},\\
    \|Pf\|_{\cC_x^s}&\le C_s\|f\|_{\cC_x^s},\\
    \|e^{\nu t\Delta}f\|_{\cC_x^s}&\le\|f\|_{\cC_x^s},
        \qquad t\ge0,\\
    \|(e^{\nu h\Delta}-I)g\|_{\cC_x^s}
        &\le C_s(\nu h)^{1/2}\|g\|_{\cC_x^{s+1}},
        \qquad h>0.
\end{align*}
The product estimate applies componentwise to tensor products. Moreover,
$t\mapsto e^{\nu t\Delta}f$ belongs to
$C([0,T];\cC_x^s)$ for every $f\in\cC_{g,x}^s$.
\end{lemma}

\begin{proof}
For $0<s<1$, the Besov norm is equivalent to
$\|f\|_\infty+[f]_s$, where $[f]_s$ is the spatial H\"older seminorm
\cite{bahouri2011fourier}. Thus
\[
    [fg]_s\le\|f\|_\infty[g]_s+[f]_s\|g\|_\infty,
\]
which gives the product estimate. The symbol of $P$ is smooth and
homogeneous of degree zero away from the origin. Its localization to a
dyadic annulus has uniformly bounded convolution-kernel $L^1$ norm, so
$\|\Delta_j^xPf\|_\infty\le C\|\Delta_j^xf\|_\infty$ on the annular
blocks. The low-frequency block contains only finitely many torus modes
and satisfies the same bound, with $P(0)$ defined to be zero.

The heat semigroup commutes with $\Delta_j^x$ and contracts $L^\infty$;
hence it contracts the stated Besov norm. For the increment estimate,
Bernstein's inequality and the semigroup contraction give
\[
    \|(e^{\nu h\Delta}-I)\Delta_j^xg\|_\infty
    \le C\min\{1,\nu h2^{2j}\}\|\Delta_j^xg\|_\infty.
\]
After multiplication by $2^{js}$, the right-hand side is at most
\[
    C\min\{2^{-j},\nu h2^j\}\|g\|_{\cC_x^{s+1}}
    \le C(\nu h)^{1/2}\|g\|_{\cC_x^{s+1}}.
\]
The finitely many low modes satisfy the same estimate. Finally, if
$f\in\cC_{g,x}^s$, choose smooth $f_n\to f$ in $\cC_x^s$. Then
\[
    \|(e^{\nu h\Delta}-I)f\|_{\cC_x^s}
    \le2\|f-f_n\|_{\cC_x^s}
       +\|(e^{\nu h\Delta}-I)f_n\|_{\cC_x^s}.
\]
First let $h\downarrow0$ and then $n\to\infty$. Continuity at other times
follows from the semigroup property.
\end{proof}

The next embedding allows us to estimate the stochastic convolution in
$C([0,T];\cC_x^s)$ by its parabolic regularity. The strict inequality
$\alpha>s$ ensures continuity in time with respect to the
$\cC_x^s$ norm.

\begin{lemma}[From parabolic regularity to spatial paths]
\label{lem:slice-embedding}
If $\alpha>s>0$, then
\begin{equation}\label{eq:slice-embedding}
    \cC^\alpha(D_T)\hookrightarrow C([0,T];\cC_x^s),
    \qquad
    \|v\|_{C([0,T];\cC_x^s)}
       \le C_{\alpha,s}\|v\|_{\cC^\alpha(D_T)}.
\end{equation}
\end{lemma}

\begin{proof}
Extend $v$ to $\R\times\T^d$ with norm bounded by twice its restriction
norm. Spatial and parabolic frequency supports imply
$\Delta_j^x\Delta_k^{\mathfrak s}=0$ unless $k\ge j-C$. Since
$\alpha>0$, summation over these parabolic blocks yields, uniformly in $t$,
\[
    \|\Delta_j^xv(t)\|_\infty
    \le C\sum_{k\ge j-C}2^{-k\alpha}\|v\|_{\cC^\alpha}
    \le C2^{-j\alpha}\|v\|_{\cC^\alpha}.
\]
This proves the norm estimate. To prove time continuity, set
$v_L:=\sum_{k\le L}\Delta_k^{\mathfrak s}v$. Each $v_L$ is smooth, and
the same calculation for the remaining blocks gives
\[
    \sup_{t\le T}\|v(t)-v_L(t)\|_{\cC_x^s}
       \le C2^{-L(\alpha-s)}\|v\|_{\cC^\alpha}.
\]
Thus $v_L\to v$ in $C([0,T];\cC_x^s)$, as claimed.
\end{proof}

\begin{lemma}[Heat smoothing with one derivative]
\label{lem:nse-smoothing}
For $0<s<1$ and $t>0$,
\begin{equation}\label{eq:nse-smoothing}
    \|e^{\nu t\Delta}P\nabla\cdot G\|_{\cC_x^s}
       \le C_s(\nu t)^{-1/2}\|G\|_{\cC_x^s}.
\end{equation}
\end{lemma}

\begin{proof}
On a spatial annulus of frequency $2^j$, the localized multiplier has
convolution-kernel $L^1$ norm bounded by
$C2^j e^{-c\nu t2^{2j}}$. Multiplying the block estimate by $2^{js}$ and
using
\[
    \sup_{r>0}r e^{-c\nu tr^2}\le C(\nu t)^{-1/2}
\]
gives the assertion on the annular blocks. The derivative annihilates the
zero mode, and the finitely many remaining low modes obey the same bound.
\end{proof}

\begin{lemma}[Bilinear estimate]\label{lem:nse-bilinear-lemma}
For $u,v\in E_\tau$, the function $N(u,v)$ belongs to $E_\tau$, vanishes
at $t=0$, and satisfies
\begin{equation}\label{eq:nse-bilinear-lemma}
    \|N(u,v)\|_{E_\tau}
       \le C_{\rm b}\sqrt{\tau/\nu}\,
          \|u\|_{E_\tau}\|v\|_{E_\tau},
\end{equation}
where $C_{\rm b}$ depends only on $d$ and $s$.
\end{lemma}

\begin{proof}
Apply Lemma~\ref{lem:nse-smoothing} and the product estimate to obtain
\begin{align*}
    \|N(u,v)(t)\|_{\cC_x^s}
    &\le C_s^2\|u\|_{E_\tau}\|v\|_{E_\tau}
       \int_0^t\big(\nu(t-\sigma)\big)^{-1/2}\,d\sigma\\
    &=2C_s^2\sqrt{t/\nu}\,
          \|u\|_{E_\tau}\|v\|_{E_\tau}.
\end{align*}
We may therefore take $C_{\rm b}=2C_s^2$. This also proves convergence to
zero at the initial time. At a positive time, split the integral a
distance $h$ from its upper endpoint. The final segment has
$\cC_x^s$ norm at most $C\sqrt{h/\nu}\|u\|_{E_\tau}\|v\|_{E_\tau}$.
On the remaining segment, the positive heat-evolution time gives an
additional spatial derivative, and the semigroup increment estimate of
Lemma~\ref{lem:nse-algebra} gives continuity. Letting $h\downarrow0$
proves continuity of the full integral. The Leray projector and the
divergence ensure that its values are divergence-free and have zero mean.
\end{proof}

\begin{proposition}[Local well-posedness and stability]
\label{prop:nse-local}
Fix $0<s<1$ and $\rho>0$. Choose $c_*>0$, depending only on $d$ and $s$,
so that $6C_{\rm b}\sqrt{c_*}\le1/2$, and set
\begin{equation}\label{eq:tau-star}
    \tau^*:=T\wedge\frac{c_*\nu}{\rho^2}.
\end{equation}
Suppose that $u_0\in\cC_{g,x}^s$ is divergence-free and mean-zero, and
that $K_\nu\phi\in E_T$ with $(K_\nu\phi)(0)=0$. If
\begin{equation}\label{eq:nse-admissible}
    \|u_0\|_{\cC_x^s}\le\rho,
    \qquad \|K_\nu\phi\|_{E_T}\le\rho,
\end{equation}
then \eqref{eq:nse-mild} has a unique solution in the closed ball
$\{u\in E_{\tau^*}:\|u\|_{E_{\tau^*}}\le3\rho\}$.
For any two data pairs satisfying these hypotheses,
\begin{equation}\label{eq:nse-stability}
    \|u^1-u^2\|_{E_{\tau^*}}
    \le2\left(
       \|u_0^1-u_0^2\|_{\cC_x^s}
       +\|K_\nu(\phi^1-\phi^2)\|_{E_T}
    \right).
\end{equation}
If $s-2<\beta_1<0$, the forcing difference satisfies
\begin{equation}\label{eq:nse-forcing-stability}
    \|K_\nu(\phi^1-\phi^2)\|_{E_T}
       \le C_{\nu,\beta_1,s,T}
            \|\phi^1-\phi^2\|_{\cC^{\beta_1}(D_T)}.
\end{equation}
\end{proposition}

\begin{proof}
\noindent\textbf{Step 1: the invariant ball.}
Set $a(t):=e^{\nu t\Delta}u_0+K_\nu\phi(t)$ and
$\mathcal T(u):=a-N(u,u)$. The strong continuity in
Lemma~\ref{lem:nse-algebra} and the assumptions on the convolution give
$a\in E_{\tau^*}$ and $\|a\|_{E_{\tau^*}}\le2\rho$.
For later reference, the bilinear estimate reads
\begin{equation}\label{eq:nse-bilinear}
    \|N(u,v)\|_{E_\tau}
       \le C_{\rm b}\sqrt{\tau/\nu}\,
          \|u\|_{E_\tau}\|v\|_{E_\tau}.
\end{equation}
Consequently, if $\|u\|_{E_{\tau^*}}\le3\rho$, then
\[
    \|\mathcal T(u)\|_{E_{\tau^*}}
       \le2\rho+9C_{\rm b}\rho^2\sqrt{\tau^*/\nu}
       \le2\rho+9C_{\rm b}\rho\sqrt{c_*}
       \le\frac{11}{4}\rho<3\rho.
\]
Thus the free term is bounded by $2\rho$, and the nonlinear term by
$3\rho/4$, so $\mathcal T$ preserves the closed ball of radius $3\rho$.

\medskip
\noindent\textbf{Step 2: contraction and the choice of time.}
Bilinearity gives
\[
    N(u,u)-N(v,v)=N(u,u-v)+N(u-v,v).
\]
On the same ball, \eqref{eq:nse-bilinear} therefore implies
\[
    \|\mathcal T(u)-\mathcal T(v)\|_{E_{\tau^*}}
       \le6C_{\rm b}\rho\sqrt{\tau^*/\nu}\,
               \|u-v\|_{E_{\tau^*}}
       \le\tfrac12\|u-v\|_{E_{\tau^*}}.
\]
The condition $\tau^*\le c_*\nu/\rho^2$ therefore gives a contraction
factor at most $1/2$, uniformly over all admissible data.
The fixed-point theorem gives existence and uniqueness in the stated
ball, and evaluation at $t=0$ gives $u(0)=u_0$.

\medskip
\noindent\textbf{Step 3: stability with respect to both inputs.}
Subtract the two mild equations. Since both solutions lie in the
$3\rho$-ball, the estimate from Step~2 yields
\begin{align*}
    \|u^1-u^2\|_{E_{\tau^*}}
    &\le\|u_0^1-u_0^2\|_{\cC_x^s}
       +\|K_\nu(\phi^1-\phi^2)\|_{E_T}\\
    &\quad+\tfrac12\|u^1-u^2\|_{E_{\tau^*}}.
\end{align*}
Absorbing the last term proves \eqref{eq:nse-stability}.
Finally, the viscous version of Lemma~\ref{lem:schauder} gives
$K_\nu:\cC^{\beta_1}(D_T)\to\cC^{\beta_1+2}(D_T)$.
Its constant may depend on $\nu$ and $T$, which are fixed. Applying
Lemma~\ref{lem:slice-embedding} with $\alpha=\beta_1+2>s$ proves
\eqref{eq:nse-forcing-stability}.
\end{proof}

For fixed $\rho$, write
\begin{equation}\label{eq:nse-local-map}
    \cS^{\mathrm{loc}}(u_0,\phi):=u\in E_{\tau^*}
\end{equation}
for this local solution. Its domain consists of the admissible data in
Proposition~\ref{prop:nse-local}; all subsequent evaluations use the same
$\rho$ and the same $\tau^*$.

\subsection{Fourier measurements and admissible coefficients}

Fix $0<s<\bar s<1$ and $M>0$, and let
\begin{equation}\label{eq:nse-initial-class}
    \cG_0^{\mathrm{NSE}}
    :=\left\{u_0\in\cC_x^{\bar s}:
       \nabla\cdot u_0=0,\ \int_{\T^d}u_0=0,
       \ \|u_0\|_{\cC_x^{\bar s}}\le M\right\}.
\end{equation}
The initial velocity is represented by its retained spatial Fourier
coefficients and reconstructed using the cutoff $\Pi^x_{J_0}$ of
Section~\ref{sec:pre-operators}. This reconstruction preserves
divergence and spatial mean; interpolation of componentwise grid values
does not in general preserve either constraint.

The measurement maps act on each of the $d$ components. Their target dimensions are
\[
    D_0:=dN_0,\qquad N_0:=|\Lambda^x_{J_0}|,
    \qquad D_J:=dN_J,\qquad N_J:=|\Lambda_J|.
\]
Thus $N_0$ and $N_J$ count scalar basis modes, while $D_0$ and $D_J$ count
real scalar measurements of a vector field. For fixed $d$,
$D_0\asymp2^{J_0d}$ and $D_J\asymp2^{J(d+2)}$.

\begin{lemma}[Spatial Fourier measurements]\label{lem:nse-spatial-encoder}
Let
\[
    c^x_{J_0}(u_0)
       :=\big(\langle u_{0,a},e_\lambda\rangle\big)_{
             1\le a\le d,\,\lambda\in\Lambda^x_{J_0}}\in\R^{D_0},
    \qquad
    (\Rec^x_{J_0}(v))_a
       :=\sum_{\lambda\in\Lambda^x_{J_0}}
                  m^x_{J_0}(\lambda)v_{a,\lambda}e_\lambda.
\]
Then $\Rec^x_{J_0}(c^x_{J_0}(u_0))=\Pi^x_{J_0}u_0$, and
\begin{align}
    \|(I-\Pi^x_{J_0})u_0\|_{\cC_x^s}
       &\le C2^{-J_0(\bar s-s)}M,
          &&u_0\in\cG_0^{\mathrm{NSE}},\label{eq:nse-initial-cutoff}\\
    \|\Rec^x_{J_0}(v)-\Rec^x_{J_0}(v')\|_{\cC_x^s}
       &\le C2^{J_0s}D_0^{1/2}|v-v'|_2.
          \label{eq:nse-initial-lip}
\end{align}
Moreover, uniformly over $J_0\ge0$,
\begin{equation}\label{eq:nse-initial-coeff}
    |c^x_{J_0}(u_0)|_\infty\le CM,
    \qquad
    \|u_0\|_{\cC_x^s}+\|\Pi^x_{J_0}u_0\|_{\cC_x^s}\le CM.
\end{equation}
Both $u_0$ and its truncation belong to $\cC_{g,x}^s$, and the truncation
is divergence-free and mean-zero.
\end{lemma}

\begin{proof}
The smooth cutoff is uniformly bounded on every spatial Besov space.
The multiplier $I-\Pi^x_{J_0}$ removes all frequencies below a fixed
multiple of $2^{J_0}$, so the remaining blocks satisfy
\[
    2^{js}\|\Delta_j^x(I-\Pi^x_{J_0})u_0\|_\infty
       \le C2^{-j(\bar s-s)}\|u_0\|_{\cC_x^{\bar s}},
       \qquad j\ge J_0-C.
\]
Taking the supremum proves \eqref{eq:nse-initial-cutoff}.
A retained trigonometric mode has frequency at most $C2^{J_0}$ and
$\cC_x^s$ norm at most $C2^{J_0s}$. The triangle inequality and
Cauchy--Schwarz over the real coefficients prove
\eqref{eq:nse-initial-lip}. Each coefficient is bounded by
$\|u_0\|_\infty\|e_\lambda\|_{L^1}$, with uniformly bounded
$\|e_\lambda\|_{L^1}$; this gives the first part of
\eqref{eq:nse-initial-coeff}. The second follows from boundedness of the
cutoffs and $\cC_x^{\bar s}\hookrightarrow\cC_x^s$.
The cutoff error also proves approximation of $u_0$ by smooth functions
in $\cC_x^s$. Finally, the cutoff acts componentwise with the same
scalar symbol, so it commutes with divergence and preserves a zero mean.
\end{proof}

For the forcing, use $c_J$ and $\Rec_J$ componentwise, now with domain
$\R^{D_J}$. Assume $s-2<\beta_1<\beta_0<0$ and write
\[
    X_0=(\cC_g^{\beta_0}(\Lambda_T))^d,
    \qquad X_1=(\cC^{\beta_1}(D_T))^d,
    \qquad \delta:=\beta_0-\beta_1>0.
\]
The estimates of Lemmas~\ref{lem:projection} and~\ref{lem:coeff} hold
componentwise. In particular,
\begin{align}
    \|(I-\Pi_J)\zeta\|_{X_1}
       &\le C2^{-J\delta}\|\zeta\|_{X_0},\label{eq:nse-noise-cutoff}\\
    |c_J(\zeta)|_\infty
       &\le C2^{-J\beta_0}\|\zeta\|_{X_0},\label{eq:nse-noise-coeff}\\
    \|\Rec_J(c)-\Rec_J(c')\|_{X_1}
       &\le CD_J^{1/2}|c-c'|_2.\label{eq:nse-noise-reconstruction}
\end{align}

The finite-dimensional domain must contain the measurement vectors and must
also allow the local solution to be evaluated at every center of the
partition of unity. Let $\mathcal V_{J_0}\subset\R^{D_0}$ and
$\mathcal W_J\subset\R^{D_J}$ be the linear subspaces defined by
\begin{align*}
    v\in\mathcal V_{J_0}
       &\quad\Longleftrightarrow\quad
       \nabla\cdot\Rec^x_{J_0}(v)=0,
       \quad\int_{\T^d}\Rec^x_{J_0}(v)=0,\\
    c\in\mathcal W_J
       &\quad\Longleftrightarrow\quad
       \nabla\cdot\Rec_J(c)=0,
       \quad\int_{\T^d}\Rec_J(c)(t,x)\,dx=0\quad\text{for all }t.
\end{align*}
These are finite linear constraints: at each nonzero spatial frequency,
the vector coefficient is orthogonal to that frequency, and the spatial
zero modes vanish.

Choose fixed constants $C_0,C_1$ sufficiently large for
\eqref{eq:nse-initial-coeff} and \eqref{eq:nse-noise-coeff}, and put
$R_J:=C_1 2^{-J\beta_0}\rho$. Define
\begin{equation}\label{eq:nse-K}
\begin{split}
    \mathcal K^{\mathrm{NSE}}_{J,J_0,\rho}
    :=\big\{(v,c)\in&\,
       ([-C_0M,C_0M]^{D_0}\cap\mathcal V_{J_0})
       \times([-R_J,R_J]^{D_J}\cap\mathcal W_J):\\
       &\|\Rec^x_{J_0}(v)\|_{\cC_x^s}\le\rho,
       \quad\|K_\nu\Rec_J(c)\|_{E_T}\le\rho\big\}.
\end{split}
\end{equation}
This set is compact: the linear constraints and the two norm constraints
are closed, and the enclosing box is compact. The reconstructed fields
are smooth, so the initial trace requirements of
Proposition~\ref{prop:nse-local} also hold. Hence every point of this set,
not only measurement vectors arising from the random forcing, defines a solution on
the common interval $[0,\tau^*]$.

\subsection{The approximation theorem}

The forcing is the centered Gaussian random variable of
Assumption~\ref{ass:noise}, interpreted componentwise, with the additional
divergence and mean constraints. The requirement $\beta_0+2>s$ ensures
that its convolution is a continuous path in $\cC_x^s$, which is stronger
than continuity of the convolution as a function of $(t,x)$.

\begin{remark}[An admissible Gaussian forcing]
One example is white-in-time forcing with a spatially smoothing
covariance. Let $(e_k)$ be a real spatial trigonometric basis without
the constant mode and set
\[
    \eta=P\sum_k\sigma_k e_k(x)\dot W_k(t),
    \qquad \sigma_k^2\asymp(1+|k|^2)^{-a},
\]
where the $\dot W_k$ are independent $\R^d$-valued temporal white
noises, realized on $\T_T$. Its parabolic regularity threshold is
$\beta_*:=\min\{-1,a-(d+2)/2\}$. More precisely, it belongs almost
surely to $\cC_g^\beta(\Lambda_T)$ for every $\beta<\beta_*$ and has no
strictly higher parabolic regularity. To see the two restrictions, a
fixed nonzero spatial mode already has the regularity of temporal white
noise, namely parabolic exponent $-1$. For the high-frequency blocks,
the pointwise variance has order
\[
    2^{2j}\sum_{0<|k|\lesssim2^j}(1+|k|^2)^{-a}
    \asymp
    \begin{cases}
       2^{j(d+2-2a)},&a<d/2,\\
       (1+j)2^{2j},&a=d/2,\\
       2^{2j},&a>d/2.
    \end{cases}
\]
Gaussian moment estimates and Bernstein's inequality, with an
arbitrarily small loss in the exponent, give the claimed regularity
below $\beta_*$. The lower variance bounds exclude any exponent above
$\beta_*$; no assertion at the endpoint is needed here. Consequently,
there is a choice $s-2<\beta_0<\beta_*$ precisely when
$a>s+(d-2)/2$. The temporal bound $-1$ remains even under arbitrarily
strong spatial smoothing.
\end{remark}

\begin{theorem}[High-probability local operator approximation]
\label{thm:nse-approx}
Let $d\ge2$, $\nu,T,M>0$, and $0<s<\bar s<1$. Suppose that $\eta$ is a
centered Gaussian Borel random variable in $X_0$, is divergence-free and
has zero spatial mean almost surely, and satisfies
\begin{equation}\label{eq:nse-beta}
    s-2<\beta_0<0.
\end{equation}
Fix $\beta_1\in(s-2,\beta_0)$ and $\delta=\beta_0-\beta_1$.
There are constants $C_0,C_\nu\ge1$ such that, for every fixed
$\rho\ge C_0M$, the following statements hold with
$\tau^*=T\wedge c_*\nu/\rho^2$.

Set
\begin{equation}\label{eq:nse-good}
    G_\rho:=\{C_\nu\|\eta\|_{X_0}\le\rho\},
    \qquad \PP(G_\rho^c)\le C e^{-c\rho^2}.
\end{equation}
For each $\varepsilon\in(0,1)$, there exist cutoffs $J,J_0$, an integer
$m$, scalar ReLU branch networks $\widetilde a_k$, and vector-valued ReLU
trunk networks $\widetilde q_k$, such that
\begin{equation}\label{eq:nse-network}
    \Gamma_\theta^{\mathrm{NSE}}(u_0,\zeta)(t,x)
       :=\sum_{k=1}^m
          \widetilde a_k\big(c^x_{J_0}(u_0),c_J(\zeta)\big)
          \widetilde q_k(t,x)
\end{equation}
satisfies the deterministic bound
\begin{equation}\label{eq:nse-uniform}
    \sup_{\substack{u_0\in\cG_0^{\mathrm{NSE}}\\
             \zeta\in X_0,\ C_\nu\|\zeta\|_{X_0}\le\rho\\
             \nabla\cdot\zeta=0,\ \int_{\T^d}\zeta=0}}
    \|\cS^{\mathrm{loc}}(u_0,\zeta)
             -\Gamma_\theta^{\mathrm{NSE}}(u_0,\zeta)\|_{Y_{\tau^*}}
       \le\varepsilon.
\end{equation}
In particular,
\begin{equation}\label{eq:nse-approx}
    \sup_{u_0\in\cG_0^{\mathrm{NSE}}}
    \E\left[\one_{G_\rho}
       \|\cS^{\mathrm{loc}}(u_0,\eta)
          -\Gamma_\theta^{\mathrm{NSE}}(u_0,\eta)\|_{Y_{\tau^*}}
       \right]\le\varepsilon,
\end{equation}
where the error inside the expectation is defined to be zero off
$G_\rho$. The numbers of real scalar measurements may be chosen to obey
\begin{equation}\label{eq:nse-counts}
    D_J\le C\left(1+\frac{\rho}{\varepsilon}\right)^{(d+2)/\delta},
    \qquad
    D_0\le C\left(1+\frac{M}{\varepsilon}\right)^{d/(\bar s-s)}.
\end{equation}
The constants are independent of $\rho$ and $\varepsilon$; those in the
probability estimate may depend on the law of $\eta$. The interval
$[0,\tau^*]$ is independent of $\varepsilon$.
\end{theorem}

\begin{proof}
We first choose an interval on which the original and truncated
solutions are defined. We then bound each of four errors by
$\varepsilon/4$: (1) forcing truncation, (2) initial-velocity truncation, (3) partition-of-unity approximation, and (4) neural approximation.

First of all, the Schauder estimate, Lemma~\ref{lem:slice-embedding}, and uniform
boundedness of the smooth cutoffs give
\begin{equation}\label{eq:nse-uniform-convolution}
    \|K_\nu\zeta\|_{E_T}
       +\sup_{J\ge0}\|K_\nu\Pi_J\zeta\|_{E_T}
       \le C_\nu\|\zeta\|_{X_0}.
\end{equation}
Indeed, the Schauder estimate yields parabolic regularity $\beta_0+2$.
Since $\beta_0+2>s$, Lemma~\ref{lem:slice-embedding} gives the bound
in $E_T$. Then, the convolution
has zero initial trace: one may approximate $\zeta\in X_0$ by smooth
forcings and pass to the limit in $E_T$.  Next, we increase $C_\nu$ so that $C_\nu\ge1$.
Also choose $C_0$ so that \eqref{eq:nse-initial-coeff} implies
\[
    \|u_0\|_{\cC_x^s}\le\rho,
    \qquad \sup_{J_0\ge0}\|\Pi^x_{J_0}u_0\|_{\cC_x^s}\le\rho
\]
whenever $u_0\in\cG_0^{\mathrm{NSE}}$ and $\rho\ge C_0M$.

For the remainder of the deterministic argument, we fix any $u_0$ and
$\zeta$ in the supremum in \eqref{eq:nse-uniform}. Then, estimate
\eqref{eq:nse-uniform-convolution} shows that $\zeta$ and all
$\Pi_J\zeta$ satisfy the convolution bound in
\eqref{eq:nse-admissible}. 

Thus Proposition~\ref{prop:nse-local} gives solutions
on the common interval $[0,\tau^*]$ satisfying
\[
\begin{aligned}
    \|\cS^{\mathrm{loc}}(u_0,\zeta)\|_{E_{\tau^*}}
        &\le 3\rho,\\
    \|\cS^{\mathrm{loc}}(u_0,\Pi_J\zeta)\|_{E_{\tau^*}}
        &\le 3\rho,\\
    \|\cS^{\mathrm{loc}}(\Pi^x_{J_0}u_0,\Pi_J\zeta)\|_{E_{\tau^*}}
        &\le 3\rho.
\end{aligned}
\]
Since these bounds are uniform in $J$ and $J_0$, the existence
interval is independent of both cutoff levels.

Next, by
\eqref{eq:nse-stability}, \eqref{eq:nse-forcing-stability}, and
\eqref{eq:nse-noise-cutoff}, we see
\begin{align}
    &\|\cS^{\mathrm{loc}}(u_0,\zeta)
             -\cS^{\mathrm{loc}}(u_0,\Pi_J\zeta)\|_{Y_{\tau^*}}
       \\
    &\quad\le2C_{\rm emb}\|K_\nu(I-\Pi_J)\zeta\|_{E_T}
       \le C\|(I-\Pi_J)\zeta\|_{X_1}
       \le C_{\rm f}2^{-J\delta}\rho.
       \label{eq:nse-forcing-error}
\end{align}
Here, $\beta_1<\beta_0$ gives a positive truncation rate, while
$\beta_1+2>s$ allows the Schauder estimate to control the error
in $E_T$. Since the forcing is unchanged, \eqref{eq:nse-stability} and the
embedding $E_{\tau^*}\hookrightarrow Y_{\tau^*}$ give the first
inequality, while \eqref{eq:nse-initial-cutoff} and
$\|u_0\|_{\cC_x^{\bar s}}\le M$ with $\bar s>s$ give the second:
\begin{align}
    &\|\cS^{\mathrm{loc}}(u_0,\Pi_J\zeta)
             -\cS^{\mathrm{loc}}(\Pi^x_{J_0}u_0,\Pi_J\zeta)
       \|_{Y_{\tau^*}}\\
    &\quad\le2C_{\rm emb}\|(I-\Pi^x_{J_0})u_0\|_{\cC_x^s}
       \le C_{\rm i}2^{-J_0(\bar s-s)}M.
       \label{eq:nse-initial-error}
\end{align}
Choose the smallest nonnegative integers $J,J_0$ satisfying
\begin{equation}\label{eq:nse-cutoff-choice}
    2^{J\delta}\ge\frac{4C_{\rm f}\rho}{\varepsilon},
    \qquad
    2^{J_0(\bar s-s)}\ge\frac{4C_{\rm i}M}{\varepsilon}.
\end{equation}
Both errors are then at most $\varepsilon/4$. The minimal choices
also give upper bounds for $J$ and $J_0$, which are needed to estimate
the numbers of measurements. Since a smallest nonnegative dyadic level satisfies
$$2^J\le2\max\{1,(4C_{\rm f}\rho/\varepsilon)^{1/\delta}\}$$ and similarly
for $J_0$, therefore we have \eqref{eq:nse-counts}.

We abbreviate $\mathcal K:=\mathcal K^{\mathrm{NSE}}_{J,J_0,\rho}$ and define
\begin{equation}\label{eq:nse-reduced-map}
    \Gamma^{\mathrm{NSE}}_{J,J_0}(v,c)
       :=\cS^{\mathrm{loc}}(\Rec^x_{J_0}(v),\Rec_J(c)),
       \qquad (v,c)\in\mathcal K.
\end{equation}
Every point of $\mathcal K$ is admissible for the local proposition, by
its definition. Moreover,
$z=(c^x_{J_0}(u_0),c_J(\zeta))$ belongs to $\mathcal K$.
The two norm constraints follow from
\eqref{eq:nse-uniform-convolution} and
\eqref{eq:nse-initial-coeff};; the coefficient bounds
follow from \eqref{eq:nse-initial-coeff} and
\eqref{eq:nse-noise-coeff}, using
$\|\zeta\|_{X_0}\le\rho/C_\nu\le\rho$. At this
measurement vector, we see
\begin{equation}\label{eq:nse-reduced-identity}
    \Gamma^{\mathrm{NSE}}_{J,J_0}(z)
       =\cS^{\mathrm{loc}}(\Pi^x_{J_0}u_0,\Pi_J\zeta).
\end{equation}

For $z=(v,c),z'=(v',c')\in\mathcal K$, stability and the reconstruction
estimates imply
\begin{align*}
    \|\Gamma^{\mathrm{NSE}}_{J,J_0}(z)
        -\Gamma^{\mathrm{NSE}}_{J,J_0}(z')\|_{Y_{\tau^*}}
    &\le C\big(2^{J_0s}D_0^{1/2}|v-v'|_2
                   +D_J^{1/2}|c-c'|_2\big)\\
    &\le L|z-z'|_2,
\end{align*}
where we take
\begin{equation}\label{eq:nse-reduced-lipschitz}
    L:=\max\left\{1,
         C\big(2^{2J_0s}D_0+D_J\big)^{1/2}\right\}.
\end{equation}

Next, we put
\begin{equation}\label{eq:nse-cover-radius}
    r:=\frac{\varepsilon}{4L},
\end{equation}
and choose an $r/2$-net $z^1,\ldots,z^m$ of $\mathcal K$ with centers
in $\mathcal K$. Compactness ensures that the net is finite. By the partition of unity Lemma~\ref{lem:pou}, we get
\[
    h_k(z):=\left(1-\frac{|z-z^k|_2}{r}\right)_+,
    \qquad
    \rho_k(z):=\frac{h_k(z)}{\sum_{j=1}^m h_j(z)},
    \qquad z\in\mathcal K.
\]
The denominator is at least $1/2$ because the centers form an
$r/2$-net. Thus $\rho_k\ge0$ and $\sum_k\rho_k=1$.
For the solutions at the partition centers,
$Q_k:=\Gamma^{\mathrm{NSE}}_{J,J_0}(z^k)$, set
\begin{equation}\label{eq:nse-pou-approximant}
    A(z)(t,x):=\sum_{k=1}^m\rho_k(z)Q_k(t,x).
\end{equation}
Each $Q_k$ is defined on the same interval, because its center
belongs to $\mathcal K$. If $\rho_k(z)>0$, then $|z-z^k|_2<r$; hence
\begin{align*}
    \|\Gamma^{\mathrm{NSE}}_{J,J_0}(z)-A(z)\|_{Y_{\tau^*}}
    &\le\sum_k\rho_k(z)
           \|\Gamma^{\mathrm{NSE}}_{J,J_0}(z)-Q_k\|_{Y_{\tau^*}}\\
    &\le\sum_k\rho_k(z)Lr
       =Lr=\frac{\varepsilon}{4}.
\end{align*}
The choice \eqref{eq:nse-cover-radius} therefore gives the required
partition-of-unity error.

Each reference solution satisfies
\[
    \|Q_k\|_{Y_{\tau^*}}\le3C_{\rm emb}\rho=:B.
\]
For $\delta_T,\delta_B>0$, choose vector-valued ReLU networks
$\widetilde q_k$ and scalar ReLU networks $\widetilde a_k$ such that
\begin{equation}\label{eq:nse-neural-tolerances}
    \|Q_k-\widetilde q_k\|_{Y_{\tau^*}}\le\delta_T,
    \qquad
    \sup_{z\in\mathcal K}|\rho_k(z)-\widetilde a_k(z)|\le\delta_B.
\end{equation}
Each partition weight $\rho_k$ is continuous on the compact set
$\mathcal K$. The universal approximation theorem
\cite{leshno1993multilayer} therefore gives a ReLU network
$\widetilde a_k$ satisfying the second inequality in
\eqref{eq:nse-neural-tolerances}.

To construct $\widetilde q_k$, let $I_hQ_k$ be the periodic multilinear
interpolant used in the proof of Theorem~\ref{thm:trunk}, applied to
each component of $Q_k$. Since $Q_k$ is uniformly continuous on
$[0,\tau^*]\times\T^d$, a sufficiently fine grid gives
\[
    \|Q_k-I_hQ_k\|_{Y_{\tau^*}}\le\frac{\delta_T}{2}.
\]
The one-dimensional basis functions have exact ReLU representations.
Approximating their products by Lemma~\ref{lem:relu-primitives} gives
a vector-valued ReLU network $\widetilde q_k$. There are only finitely
many interpolation terms, so the multiplication accuracies can be
chosen such that
\[
    \|I_hQ_k-\widetilde q_k\|_{Y_{\tau^*}}\le\frac{\delta_T}{2}.
\]
The triangle inequality proves the first bound in
\eqref{eq:nse-neural-tolerances}. The network depends on the spatial
variables through the periodic one-dimensional basis functions, so its
values agree on opposite faces of $[0,2\pi]^d$. Thus
$\widetilde q_k$ defines an element of $Y_{\tau^*}$.

Define $\widetilde A(z):=\sum_k\widetilde a_k(z)\widetilde q_k$.
Adding and subtracting $\sum_k\rho_k(z)\widetilde q_k$ gives
\begin{align}
    \|A(z)-\widetilde A(z)\|_{Y_{\tau^*}}
    &\le\sum_k\rho_k(z)\|Q_k-\widetilde q_k\|_{Y_{\tau^*}}
        \\
    &\quad+\sum_k|\rho_k(z)-\widetilde a_k(z)|
                      \|\widetilde q_k\|_{Y_{\tau^*}}\\
    &\le\delta_T+m(B+\delta_T)\delta_B.
       \label{eq:nse-neural-error}
\end{align}
The partition property bounds the first sum by $\delta_T$
independently of $m$. The second sum contains $m$ errors, each
bounded by $(B+\delta_T)\delta_B$. We therefore choose
\begin{equation}\label{eq:nse-neural-choice}
    \delta_T:=\frac{\varepsilon}{8},
    \qquad
    \delta_B:=\frac{\varepsilon}{8m(B+1)}.
\end{equation}
As $\varepsilon<1$, these choices give
$\delta_T+m(B+\delta_T)\delta_B\le\varepsilon/4$.
The network in \eqref{eq:nse-network} is
$\widetilde A(c^x_{J_0}(u_0),c_J(\zeta))$.

With $z=(c^x_{J_0}(u_0),c_J(\zeta))$, we use
\eqref{eq:nse-reduced-identity} to write
\begin{align}\label{e.w10281}
    &\|\cS^{\mathrm{loc}}(u_0,\zeta)
                -\Gamma_\theta^{\mathrm{NSE}}(u_0,\zeta)\|_{Y_{\tau^*}}\\
    &\quad\le
       \|\cS^{\mathrm{loc}}(u_0,\zeta)
                 -\cS^{\mathrm{loc}}(u_0,\Pi_J\zeta)\|_{Y_{\tau^*}}\\
    &\qquad+
       \|\cS^{\mathrm{loc}}(u_0,\Pi_J\zeta)
                 -\cS^{\mathrm{loc}}(\Pi^x_{J_0}u_0,\Pi_J\zeta)
          \|_{Y_{\tau^*}}\\
    &\qquad+
       \|\Gamma^{\mathrm{NSE}}_{J,J_0}(z)-A(z)\|_{Y_{\tau^*}}
       +\|A(z)-\widetilde A(z)\|_{Y_{\tau^*}}\\
    &\quad\le\frac{\varepsilon}{4}+\frac{\varepsilon}{4}
              +\frac{\varepsilon}{4}+\frac{\varepsilon}{4}
       =\varepsilon.
\end{align}
Every bound is uniform over the initial-data class and the deterministic
forcing ball in \eqref{eq:nse-uniform}. This proves that estimate with
one network for all these inputs.

Lastly, Fernique's estimate in Lemma~\ref{lem:fernique}, applied to the
vector-valued Gaussian variable, gives
\[
    \PP(G_\rho^c)
       =\PP(\|\eta\|_{X_0}>\rho/C_\nu)
       \le C\exp(-c\rho^2/C_\nu^2)
       \le C e^{-c'\rho^2}.
\]
Fix $u_0\in\cG_0^{\mathrm{NSE}}$ and define
\[
    Z_{u_0}:=
    \begin{cases}
    \|\cS^{\mathrm{loc}}(u_0,\eta)
       -\Gamma_\theta^{\mathrm{NSE}}(u_0,\eta)\|_{Y_{\tau^*}},
       &\text{on }G_\rho,\\
    0,&\text{on }G_\rho^c.
    \end{cases}
\]
By \eqref{eq:nse-stability}, \eqref{eq:nse-forcing-stability}, and
the continuity of restriction from $X_0$ to $X_1$, the solution map
is continuous on the admissible forcing set. The neural approximation
is also continuous in the forcing, since the Fourier measurements and
ReLU networks are continuous. Thus $Z_{u_0}$ is measurable.

The uniform estimate \eqref{eq:nse-uniform} gives
$Z_{u_0}\le\varepsilon\one_{G_\rho}$ almost surely, and hence
\[
    \E Z_{u_0}\le\varepsilon\PP(G_\rho)\le\varepsilon.
\]
This bound is uniform in $u_0\in\cG_0^{\mathrm{NSE}}$, which proves
\eqref{eq:nse-approx}.

\end{proof}

\bibliographystyle{plain}
\bibliography{ref0909}

\end{document}